\documentclass{article}
\usepackage[dvipsnames]{xcolor}

\usepackage[nonatbib, final]{neurips_2021} %[1] format

\usepackage{cite} %allows [1-3] instead of [1,2,3]

\usepackage[utf8]{inputenc} % allow utf-8 input
\usepackage[T1]{fontenc}    % use 8-bit T1 fonts
\usepackage[colorlinks=true, citecolor=blue, linkcolor=black, urlcolor=black]{hyperref}
\usepackage{url}            % simple URL typesetting
\usepackage{booktabs}       % professional-quality tables
\usepackage{amsfonts}       % blackboard math symbols
\usepackage{nicefrac}       % compact symbols for 1/2, etc.

\usepackage{microtype}      % microtypography
\usepackage{xcolor}         % colors

\usepackage{amsmath,amsfonts,bm}

\def\eqref#1{equation~\ref{#1}}
\def\ceil#1{\lceil #1 \rceil}

\def\1{\bm{1}}

\def\vone{{\bm{1}}}
\def\vmu{{\bm{\mu}}}
\def\vtheta{{\bm{\theta}}}

\def\vg{{\bm{g}}}

\def\mA{{\bm{A}}}
\def\mB{{\bm{B}}}

\def\mD{{\bm{D}}}

\def\mZ{{\bm{Z}}}

\def\mSigma{{\bm{\Sigma}}}

\DeclareMathAlphabet{\mathsfit}{\encodingdefault}{\sfdefault}{m}{sl}
\SetMathAlphabet{\mathsfit}{bold}{\encodingdefault}{\sfdefault}{bx}{n}

\DeclareMathOperator*{\argmin}{arg\,min}

\usepackage{graphicx}       % -----------
\usepackage{amsmath,amsthm,amssymb, systeme, bbold}
\usepackage{mathtools}
\usepackage{xspace}
\usepackage{wrapfig} %to wrap figures, tables, algorithms
\usepackage{algorithm}
\usepackage{algorithmic}
\usepackage{comment}

\usepackage{makecell} % two lines in one table cell
\usepackage{xcolor}
\usepackage{tikz} % add text to figures
\usepackage{caption} %\captionsetup[table]{skip=10pt}
\usepackage{dblfloatfix} %enable bottom of the page figure

\makeatletter
\newcommand{\mypm}{\mathbin{\mathpalette\@mypm\relax}}
\newcommand{\@mypm}[2]{\ooalign{%
  \raisebox{.1\height}{$#1+$}\cr
  \smash{\raisebox{-.6\height}{$#1-$}}\cr}}
\makeatother
\DeclareMathOperator{\sgn}{sgn}

\newcommand{\bo}[1]{\boldsymbol{#1}}

\newcommand{\myqed}{\tag*{$\square$}}

\theoremstyle{definition}

\newtheorem{theorem}{Theorem}[section]

\theoremstyle{remark}

\title{Gradient-based Hyperparameter Optimization \\Over Long Horizons}

\author{Paul Micaelli \\
University of Edinburgh\\
\texttt{\{paul.micaelli\}@ed.ac.uk} \\
\And
Amos Storkey \\
University of Edinburgh\\
\texttt{\{a.storkey\}@ed.ac.uk}}

\begin{document}

\maketitle

\begin{abstract}
Gradient-based hyperparameter optimization has earned a widespread popularity in the context of few-shot meta-learning, but remains broadly impractical for tasks with long horizons (many gradient steps), due to memory scaling and gradient degradation issues. A common workaround is to learn hyperparameters online, but this introduces greediness which comes with a significant performance drop. We propose forward-mode differentiation with sharing (FDS), a simple and efficient algorithm which tackles memory scaling issues with forward-mode differentiation, and gradient degradation issues by sharing hyperparameters that are contiguous in time. We provide theoretical guarantees about the noise reduction properties of our algorithm, and demonstrate its efficiency empirically by differentiating through $\sim 10^4$ gradient steps of unrolled optimization. We consider large hyperparameter search ranges on CIFAR-10 where we significantly outperform greedy gradient-based alternatives, while achieving $\times 20$ speedups compared to the state-of-the-art black-box methods. Code is available at: \url{https://github.com/polo5/FDS}
%Code will be available upon publication. TODO-> submit code
\end{abstract}

\section{Introduction}
Deep neural networks have shown tremendous success on a wide range of applications, including classification \cite{He2015ResNet}, generative models \cite{Bock2018BigGan}, natural language processing \cite{Devlin2018BERT} and speech recognition \cite{VanDenOord2018WaveNet}. This success is in part due to effective optimizers such as SGD with momentum or Adam \cite{Kingma2015Adam}, which require carefully tuned hyperparameters for each application. In recent years, a long list of heuristics to tune such hyperparameters has been compiled by the deep learning community, including things like: how to best decay the learning rate \cite{Loshchilov2017WarmRestarts}, how to scale hyperparameters with the budget available \cite{Li2020BudgetedSchedules}, and how to scale learning rate with batch size \cite{Goyal2017LargeMinibatchSGD}. Unfortunately these heuristics are often dataset specific and architecture dependent \cite{Dong2020AutoHAS}. They also don't apply well to new optimizers \cite{loshchilov2018AdamW}, or new tools, like batch normalization which allows for larger learning rates \cite{Ioffe2015BatchNorm}.

With so many ways to choose hyperparameters, the deep learning community is at risk of adopting models based on how much effort went into tuning them, rather than their methodological insight. The field of hyperparameter optimization (HPO) aims to find hyperparameters that provide a good generalization performance automatically. Unfortunately, existing tools are rather unpopular for deep learning, largely owing to their computational cost. The method developed here is a gradient-based HPO approach; that is, it calculates hypergradients, i.e. the gradient of some validation loss with respect to each hyperparameter. Gradient-based HPO should be more efficient than black-box methods as the dimensionality of the hyperparameter space increases, since it is able to utilize gradient information rather than rely on trial and error. In practice however, learning hyperparameters with gradients has only been popular in few-shot learning tasks where the horizon is short, i.e. where the underlying task is solved with a few gradient steps. This is because long horizons cause hypergradient degradation, and incur a memory cost that makes reverse-mode differentiation prohibitive.

\begin{wrapfigure}[26]{t}{0.5\textwidth}
\begin{minipage}[t]{0.5\textwidth}
%\vspace{-0.5cm}
\vspace{-0.1cm}

%\begin{figure*}[t!] %!htbp
\centering
\includegraphics[width=\textwidth]{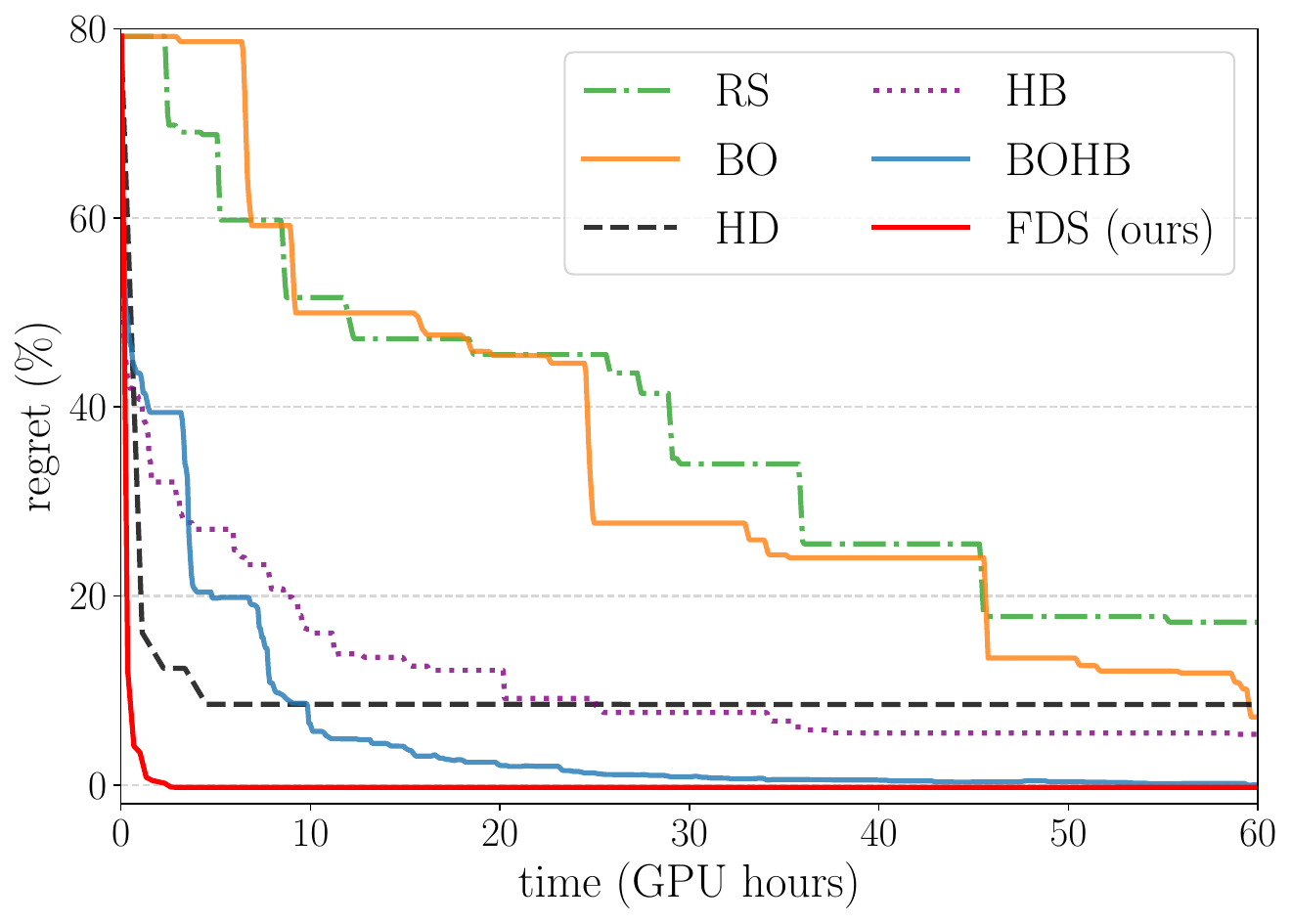}
\caption{Performance of the most popular hyperparameter optimization methods when learning the learning rate, momentum and weight decay on $50$ epochs of CIFAR-10, for a WideResNet of 16 layers. Non-greedy methods (RS, BO, HB\cite{Li2017Hyperband}, BOHB\cite{Falkner2018BOHB}) solve for global hyperparameters but rely on trial-and-error which makes them slow. Greedy gradient-based methods (e.g. HD \cite{Baydin2018HypergradientDescent}) are faster but solve for local hyperparameters which makes them suboptimal. Our method combines the strengths of these two paradigms and outperforms the next best method while converging $20$ times faster. Each curve is the average of $8$ seeds.}
\label{fig:perf over time}
% \end{figure*}
\end{minipage}
\end{wrapfigure}

\noindent Greedy gradient-based methods alleviate both of these issues by calculating local hypergradients based on intermediate validation losses. Unfortunately, this introduces some bias \cite{Wu2018ShortHorizonBias} and results in a significant performance drop, which we are able to quantify in this work. We make use of forward-mode differentiation, which has been shown to offer a memory cost constant with horizon size. However, previous forward-mode methods don't address gradient degradation explicitly and are thus limited to the greedy setting \cite{Franceschi2017ForwardAndReverseGradientBasedHO,Donini2019SchedulingLearningRateNewInsights}.

We introduce FDS (Foward-mode Differentiation with hyperparameter Sharing), which to the best of our knowledge demonstrates for the first time that hyperparameters can be differentiated non-greedily over long horizons. Specifically, we make the following contributions: \textbf{(1)} we propose to share hyperparameters through time, both motivating it theoretically and with various experiments, \textbf{(2)} we combine the above in a forward-mode differentiation algorithm, and \textbf{(3)} we show that our method can significantly outperform various HPO algorithms, for instance when learning the hyperparameters of the SGD-momentum optimizer.

\section{Related Work}

There are many ways to perform hyperparameter optimization (HPO) \cite{Feurer2019HyperparamOptBookChapter}, including Bayesian optimization (BO) \cite{Snoek2015ScalableBayesianOptimizationDeepNets}, reinforcement learning \cite{Zoph2017NASwithRL}, evolutionary algorithms \cite{Jaderberg2017PopBasedTrainingNets} and gradient-based methods \cite{Bengio2000GradientBasedHyperparamOpt}. The state-of-the-art in HPO depends on the problem setting, but black-box methods like Hyperband (HB) \cite{Li2017Hyperband}, and its combination with BO into a method called BOHB \cite{Falkner2018BOHB} have been the most popular. Modern work in meta-learning deals with various forms of gradient-based HPO \cite{Hospedales2020MetaLearningSurvey}, but usually focuses on the few-shot regime \cite{Finn2017MAML} where horizons are conveniently short ($\sim 10$ steps) while we focus on long horizons ($\sim 10^4$ steps).

\paragraph{Gradient-based HPO.} Using the gradient of some validation loss with respect to the hyperparameters is typically the preferred choice when the underlying optimization is differentiable. This is a type of constrained optimization \cite{Franceschi2018BilevelPF} which stems from earlier work on backpropagation through time \cite{Werbos1990BackpropagationThroughTime} and real-time recurrent learning \cite{Williams1989AlgorithmRecurrentNeuralNetwork}. Unfortunately, differentiating optimization is an expensive procedure in both time and memory, and most proposed methods are limited to small models and toy datasets \cite{Domke2012OptimizationBasedModeling, Maclaurin2015GradientHO, Pedregosa2016GradientHO}. Efforts to make the problem more tractable include optimization shortcuts \cite{Fu2016DrMAD}, truncation \cite{Shaban2018TruncatedBackPropBilevelOpt} and implicit gradients \cite{Larsen1996OptimalUseOfValidation, Rajeswaran2019ImplicitMAML, Lorraine2019MillionsHyperparamOptimImplicit}. Truncation can be combined with our approach but produces biased gradients \cite{Metz2019PathologiesInTrainingOptimizers}, while implicit differentiation is only applicable to hyperparameters that define the training loss (e.g. augmentation) but not to hyperparameters that define how the training loss is minimized (e.g. optimizer hyperparameters). Forward-mode differentiation \cite{Williams1989AlgorithmRecurrentNeuralNetwork} boasts a memory cost constant with horizon size, but gradient degradation has restricted its use to the greedy setting \cite{Franceschi2017ForwardAndReverseGradientBasedHO,Donini2019SchedulingLearningRateNewInsights}.

\paragraph{Greedy Methods.} One trick that prevents gradient degradation and significantly reduces compute and memory cost is to solve the bilevel optimization greedily. This has become the default trick in various long-horizon problems, including HPO over optimizers \cite{Luketina2016GradBasedRegLearning, Franceschi2017ForwardAndReverseGradientBasedHO, Baydin2018HypergradientDescent, Donini2019SchedulingLearningRateNewInsights}, architecture search \cite{Liu2019DARTS}, dataset distillation \cite{Wang2018DatasetDistillation} or curriculum learning \cite{Ren2018LearningToReweight}. Greediness refers to finding the best hyperparameters locally rather than globally. In practice, it involves splitting the inner optimization problem into smaller chunks (often just one batch), and solving for hyperparameters over these smaller horizons instead; often in an online fashion. We formalize greediness in Section \ref{sec:greediness}. In this paper we expand upon previous observations \cite{Wu2018ShortHorizonBias} and take the view that greediness fundamentally solves for the wrong objective. Instead, the focus of our paper is to extend forward-mode differentiation methods to the non-greedy setting.

\paragraph{Gradient Degradation.} Gradient degradation of some scalar w.r.t a parameter is a broad issue that arises when that parameter influences the scalar in a chaotic fashion, such as through long chains of nonlinear mappings. This manifests itself in HPO as vanishing or exploding hypergradients, due to low or high curvature components of the validation loss surface. This leads to hypergradients that have a large variance, making differentiation through long horizons impractical. This is usually observed in the context of recurrent neural networks \cite{Bengio1993RNNGradientIssues, Bengio1994LongTermGradInefficient}, but also in reinforcement learning \cite{Parmas2018PIPPSCurseOfChaos} and HPO \cite{Maclaurin2015GradientHO}. Solutions like LSTMs \cite{Hochreiter1997LSTM} and gradient clipping \cite{Pascanu2013DifficultyOfRNNs} have been proposed, but are respectively inapplicable and insufficient to long-horizon HPO. Variational optimization \cite{Metz2019PathologiesInTrainingOptimizers} and preconditioning warp-layers \cite{Flennerhag2020WarpedGradientDescent} can mitigate gradient degradation, but these methods are expensive in memory and therefore are limited to small architectures and/or a few hundred inner steps. In comparison, we differentiate over $\sim 10^4$ inner steps for WideResNets \cite{Zagoruyko2016WRN}.

\section{Background}

\subsection{Problem statement}
\label{sec:problem statement}

Consider a neural network with weights $\bo{\theta}$, trained to minimize a loss $\mathcal{L}_{\text{train}}$ over a dataset $\mathcal{D}_{\text{train}}$. This is done by taking $T$ steps with a gradient-based optimizer $\Phi$, which uses a collection of hyperparameters $\bo{\lambda} \in \mathbb{R}^{T}$. For clarity of notation, consider that $\Phi$ uses one hyperparameter per step, written $\bo{\lambda}_{[t]}$, where indices are shown in brackets to differentiate them from a variable evaluated at time $t$. We can explicitly write out this optimizer as $\Phi : \bo{\theta}_{t+1} = \Phi(\mathcal{L}_{\text{train}}(\bo{\theta}_{t}(\bo{\lambda}_{[1:t]}), \mathcal{D}_{\text{train}}), \bo{\lambda}_{[t+1]}) ~~ \forall t \in \{1,2,\ldots,T\}$. Note that $\bo{\theta}_t$ is a function of $\bo{\lambda}_{[1:t]}$, and it follows that $\bo{\theta}_T = \bo{\theta}_T(\bo{\lambda}_{[1:T]}) = \bo{\theta}_T(\bo{\lambda})$.

We would like to find the hyperparameters $\bo{\lambda}^*$ such that the result, at time $T$, of the gradient process minimizing objective $\mathcal{L}_{\text{train}}$, also minimizes some generalization loss $\mathcal{L}_{\text{val}}$ on a validation dataset $\mathcal{D}_{\text{val}}$. This can be cast as the following constrained optimization:
% \begin{gather}
% OLD VERSION:\\
% \hspace*{-3cm} \bo{\lambda}^* = \argmin_{\bo{\lambda}} \mathcal{L}_{\text{val}}(\bo{\theta}_T(\bo{\lambda}), \mathcal{D}_{\text{val}}) \label{eq:outer_loop}
% \\ \hspace*{-0cm} \text{subject to} ~~~ \bo{\theta}_T = \argmin_{\bo{\theta}} \mathcal{L}_{\text{train}}(\bo{\theta}, \mathcal{D}_{\text{train}}) 
% \label{eq:inner_loop}
% \\ \hspace*{+1cm} \text{solved with} ~~~ \bo{\theta}_{t+1} = \Phi(\bo{\theta}_t(\bo{\lambda}_{[0:t-1]}),\bo{\lambda}_{[t]})
% \label{eq:bilevel_solved_with}
% \end{gather}
\begin{gather}
\hspace*{-5cm} \bo{\lambda}^* = \argmin_{\bo{\lambda}} \mathcal{L}_{\text{val}}(\bo{\theta}_T(\bo{\lambda}), \mathcal{D}_{\text{val}}) \label{eq:outer_loop}
\\ \hspace*{-0cm} \text{subject to}  ~~~ \bo{\theta}_{t+1} = \Phi(\mathcal{L}_{\text{train}}(\bo{\theta}_{t}(\bo{\lambda}_{[1:t]}), \mathcal{D}_{\text{train}}), \bo{\lambda}_{[t+1]})\
\label{eq:inner_loop}
\end{gather}
%~~ \forall t=1,\ldots,T
% \\ \hspace*{-0cm} \text{where } \Phi \text{ is chosen to target} \nonumber \\
% \argmin_{\bo{\theta}} \mathcal{L}_{\text{train}}(\bo{\theta}, \mathcal{D}_{\text{train}}) 

The inner loop in Eq \ref{eq:inner_loop} expresses a constraint on the outer loop in Eq \ref{eq:outer_loop}. In gradient-based HPO, our task is to compute the hypergradient $d\mathcal{L}_{\text{val}} / d\bo{\lambda}$ and update $\bo{\lambda}$ accordingly. Note that some methods require $\bo{\theta}_T = \argmin_{\bo{\theta}} \mathcal{L}_{\text{train}}(\bo{\theta}, \mathcal{D}_{\text{train}}, \bo{\lambda})$ with $\mathcal{L}_{\text{train}}$ being a function of $\bo{\lambda}$ explicitly \cite{Larsen1996OptimalUseOfValidation, Rajeswaran2019ImplicitMAML, Lorraine2019MillionsHyperparamOptimImplicit}, in which case the above becomes a bilevel optimization \cite{Stackelberg1952BilevelOptimization}. Our algorithm waives these requirements.

\subsection{Greediness}
\label{sec:greediness}

Let $H$ be the horizon, which corresponds to the number of gradient steps taken in the inner loop (to optimize $\bo{\theta}$) before one gradient step is taken in the outer loop (to optimize $\bo{\lambda}$). When solving Eq \ref{eq:outer_loop} non-greedily we have $H=T$. However, most modern approaches are greedy \cite{Luketina2016GradBasedRegLearning, Franceschi2017ForwardAndReverseGradientBasedHO, Baydin2018HypergradientDescent,Liu2019DARTS,Wang2018DatasetDistillation}, in that they rephrase the above problem into a sequence of several independent problems of smaller horizons, where $\bo{\lambda}^*_{[t:t+H]}$ is learned in the outer loop subject to an inner loop optimization from $\bo{\theta}_t$ to $\bo{\theta}_{t+H}$ with $H \ll T$. Online methods such as Hypergradient Descent (HD) \cite{Baydin2018HypergradientDescent} are an example of greedy differentiation, where $H=1$ and $\bo{\lambda}_{[t]}$ is updated at time $t$. Instead, non-greedy algorithms like FDS only update $\bo{\lambda}_{[t]}$ at time $T$.

Greediness has many advantages: it mitigates gradient degradation, makes hypergradients cheaper to compute, and it requires less memory when used with reverse-mode differentiation. However, greedy methods look for $\bo{\lambda}^*$ that does well locally rather than globally, i.e. they constrain $\bo{\lambda}^*$ to a subspace of solutions such that $\bo{\theta}_{H}, \bo{\theta}_{2H}, ..., \bo{\theta}_{T}$ all yield good validation performances. In our experiments, we found that getting competitive hyperparameters with greediness often revolves around tricks like online learning with a very low outer learning rate combined with hand-tuned initial hyperparameter values, to manually prevent convergence to small values. But solving the greedy objective correctly leads to poor solutions, a special case of which was previously described as the ``short-horizon bias'' \cite{Wu2018ShortHorizonBias}. In FDS, we learn global hyperparameters in a non-greedy fashion, which means that the hypergradient  $d\mathcal{L}_{\text{val}} / d\bo{\lambda}$ is only ever calculated at $\bo{\theta}_{T}$.

\subsection{Forward-mode differentiation of modern optimizers}
\label{sec: forward mode diff background}
The vast majority of meta-learning applications use reverse-mode differentiation in the inner optimization problem (Eq \ref{eq:inner_loop}) to optimize $\bo{\theta}$. However, the memory cost of using reverse-mode differentiation for the outer optimization (Eq \ref{eq:outer_loop}) is $\mathcal{O}(FH)$, where $F$ is the memory used by one forward pass through the network (weights plus activations). This is often referred to as backpropagation through time (BPTT). In the non-greedy setting where $H=T$, BPTT is extremely limiting: for large networks, only $T \sim 10$ gradient steps could be solved with modern GPUs, while problems like CIFAR-10 require $T \sim 10^4$. Instead, we make use of forward-mode differentiation, which scales in memory as $\mathcal{O}(DN)$, where $D$ is the number of weights and $N$ is the number of learnable hyperparameters. The additional scaling with $N$ is a limitation if we learn one hyperparameter per inner step ($N = T$). Sharing hyperparameters (section \ref{subsec:sharing_hyperparameters}) mitigates gradient degradation, but also conveniently allows for smaller values of $N$.

For clarity of notation, we consider forward-mode hypergradients for the general case of using one hyperparameter per step, i.e. $\bo{\lambda} \in \mathbb{R}^{T}$. First, we use the chain rule to write $d\mathcal{L}_{\text{val}}/ d\bo{\lambda} = (\partial\mathcal{L}_{\text{val}}/ \partial\bo{\theta}_T) (d\bo{\theta}_T/d\bo{\lambda})$ where the direct gradient has been dropped since $\partial \mathcal{L}_{\text{val}}/\partial \bo{\lambda} = 0$ for optimizer hyperparameters. The first term on the RHS is trivial and can be obtained with reverse-mode differentiation as usual. The second term is more problematic because $\bo{\theta}_T = \bo{\theta}_T(\bo{\theta}_{T-1}(\bo{\theta}_{T-2}(...), \bo{\lambda}_{[T-1]}), \bo{\lambda}_{[T]}) $. We use the chain rule again to calculate this term recursively:
\begin{equation}
\dfrac{d\bo{\theta}_t}{d\bo{\lambda}} = \dfrac{\partial\bo{\theta}_t}{\partial\bo{\theta}_{t-1}}\biggr\rvert_{\bo{\lambda}} \dfrac{d\bo{\theta}_{t-1}}{d\bo{\lambda}} + \dfrac{\partial\bo{\theta}_t}{\partial\bo{\lambda}}\biggr\rvert_{\bo{\theta}_{t-1}} ~~~ \text{which we write as}~~~ \bo{Z}_t = \bo{A}_t \bo{Z}_{t-1} + \bo{B}_t
\end{equation}
where $\bo{\theta}_t \in \mathbb{R}^{D}$, $\bo{\lambda} \in \mathbb{R}^{T}$, $\bo{Z}_t \in \mathbb{R}^{D \times T}$, $\bo{A}_t \in \mathbb{R}^{D \times D}$ and $\bo{B}_t \in \mathbb{R}^{D \times T}$. Note that the columns of $\bo{Z}$ at step $t$ are zeros for indices $t+1, t+2, ..., T$, since the hyperparameters at those steps haven't been used yet. 

The expressions for $\bo{A}_t$ and $\bo{B}_t$ depend on the specific hyperparameters we are differentiating. In this work, we consider the most popular optimizer, namely SGD with momentum and weight decay. To the best of our knowledge, previous work focuses on simpler versions of this optimizer, usually by removing momentum and weight decay, and only learns the learning rate, greedily. We use Pytorch's update rule for SGD \cite{Paszke2019Pytorch}, namely $\bo{\theta}_t = \Phi(\bo{\theta}_{t-1}) = \bo{\theta}_{t-1} - \alpha_t \bo{v}_t $ with learning rate $\alpha_t$, momentum $\beta_t$, weight decay $\xi_t$ and velocity $\bo{v}_t = \beta_t \bo{v}_{t-1} + (\partial \mathcal{L}_{\text{train}}/\partial \bo{\theta}_{t-1}) + \xi_t \bo{\theta}_{t-1} $. Consider the case when we learn the learning rate schedule, namely $\bo{\lambda} = \bo{\alpha}$. If we use the update rule without momentum \cite{Donini2019SchedulingLearningRateNewInsights}, $\bo{B}_t$ is conveniently sparse: it is a $D \times T$ matrix that only has one non-zero column at index $t$ corresponding to $\partial\bo{\theta}_t /\partial\alpha_t$. However, we include terms like momentum and therefore the velocity depends on the hyperparameters of previous steps. In that case, a further recursive term $\bo{C}_t = (\partial \bo{v}_t / \partial \bo{\lambda})$ must be considered to get exact hypergradients. Putting it together (see Appendix A) we obtain:
\begin{equation}
\systeme{
\bo{A}_t^{\bo{\alpha}} = \vone  - \alpha_t \left(\dfrac{\partial^2 \mathcal{L}_{\text{train}}}{\partial \bo{\theta}_{t-1}^2} + \xi_t \vone\right),
\bo{B}_t^{\bo{\alpha}} = -\beta_t \alpha_t \bo{C}_{t-1}^{\bo{\alpha}} - \delta_t^{\otimes}\left(\beta_t \bo{v}_{t-1} + \dfrac{\partial \mathcal{L}_{\text{train}}}{\partial \bo{\theta}_{t-1}} + \xi_t \bo{\theta}_{t-1}\right),
\bo{C}_t^{\bo{\alpha}} = \beta_t \bo{C}_{t-1}^{\bo{\alpha}} + \left(\xi_t \vone + \dfrac{\partial^2 \mathcal{L}_{\text{train}}}{\partial \bo{\theta}_{t-1}^2} \right) \bo{Z}_{t-1}^{\bo{\alpha}}
}
\label{eq:recursive_alpha}
\end{equation}
where $\vone$ is a $D\times D$ identity matrix, and $\delta_t^{\otimes}(\bo{q})$ turns a vector $\bo{q}$ of size $D$ into a matrix of size $D \times T$, whose $t$-th column is set to $\bo{q}$ and other columns to $\bo{0}$s. While the matrices in Eq \ref{eq:recursive_alpha} are updated online, the hyperparameters aren't. This is because in the non-greedy setting we don't have access to the hypergradients until we have computed $\bo{Z}_T^{\bo{\alpha}}$. A similar technique can be applied to momentum and weight decay to get $\bo{Z}_T^{\bo{\beta}}$ and $\bo{Z}_T^{\bo{\xi}}$ (see Appendix A). All hypergradient derivations in this paper were checked with finite differences. Note that we focus on learning the hyperparameters of SGD with momentum  because it is the most common optimizer in the deep learning literature. However, just like reverse-mode differentiation, forward-mode differentiation can be applied to any differentiable hyperparameter, albeit with appropriate $\bo{A_t}$ and $\bo{B_t}$ matrices.

\section{Non-greedy Differentiation Over Long Horizons}
\label{sec:enabling_nongreediness}

\subsection{Hyperparameter sharing: trading off noise reduction with bias increase}
\label{subsec:sharing_hyperparameters}

The main challenge of doing non-greedy meta-learning over long horizons is gradient degradation. In HPO this arises because a small change in $\mathcal{L}_{\text{train}}(\bo{\theta}_t)$ can cascade forward into a completely different $\bo{\theta_T}$, resulting in large fluctuations of the hypergradients. This noise makes the generalization loss hard to minimize (Eq \ref{eq:outer_loop}). We find that the two main causes for this noise are the ordering of the training minibatches, and the weight initialization $\bo{\theta}_0$. Ideally, the hyperparameters we learn should be agnostic to both of these factors, and so we would like to average out their effect on hypergradients. 

\paragraph{Ensemble averaging} The most obvious way to address the above is to compute all hypergradients across several random seeds, where each seed corresponds to a different dataset ordering and weight initialization. We could then obtain an average hypergradient $\mu_t$ for each inner step $t$. Here, $\vmu \in \mathbb{R}^T$ is often called an ensemble average in statistical mechanics. The issue with ensemble averaging in our setting is its computational and memory cost, since each random seed requires differentiating through $T$ unrolled inner steps for $T$ hyperparameters. This makes both reverse-mode and forward-mode differentiation intractable. We consider the ensemble average as optimal in our analysis, which allows us to derive an expression for the mean square error between our hypergradients estimate and $\vmu$.

\paragraph{Time averaging} In FDS, we use the long horizon to our advantage and propose to do time averaging instead of ensemble averaging, i.e. we \emph{average out hypergradients across the inner loop (Eq \ref{eq:inner_loop}) rather than the outer loop (Eq \ref{eq:outer_loop})}. More specifically, we sum hypergradients from $W$ neighbouring time steps in the inner loop, which is exactly equivalent to sharing one hyperparameter over all these steps. The average is then obtained trivially by dividing by $W$. For instance, when learning the learning rate schedule $\bo{\alpha}$ for $H = 10^4$ inner steps, we can learn $\bo{\alpha} \in \mathbb{R}^{10}$ where each learning rate is used for a window of $W=10^3$ contiguous gradient steps. A stochastic system where the time average is equivalent to the ensemble average is called \emph{ergodic} \cite{Walters1982ErgodicTheory} and has been the subject of much research in thermodynamics \cite{Boltzmann1896Ergodicity} and finance \cite{Peters2019ErgodicityEconomics}. In our case, hypergradients aren't generally ergodic, and so using a single average hypergradient for $W$ contiguous steps can introduce a bias. Informally, time averaging contiguous hypergradients leads to both noise reduction and bias increase, and we flesh out the nature of this trade-off in Theorem \ref{theorem}.
%\smallskip
\begin{theorem} Let each time step $t \in \{1, 2, ..., T\}$ have a corresponding hyperparameter $\lambda_t$ and non-greedy hypergradient $g_t=\partial L_{\text{val}}(\vtheta_T)/ \partial \lambda_t$. Each $g_t$ is viewed as a random variable due to the sampling process of the weight initialization $\vtheta_0$ and the inner loop minibatch selection. Let $\vg = [g_1,g_2,\ldots, g_{T}]$ be sufficiently well approximated by a Gaussian distribution $\vg\sim \mathcal{N}(\vmu,\mSigma)$, with mean $\vmu=[\mu_1, \mu_2, ..., \mu_T]$ and covariance matrix $\mSigma$, where $\vmu$ corresponds to the optimal hypergradients. Assume that the changes in the values of $\vmu$ over time are bounded, i.e. $\vmu$ can be written as the $\epsilon$-Lipschitz function $\mu_{t+1} = \mu_t + \epsilon_t$, where $\epsilon_t \in [-\epsilon,\epsilon]$. Finally, let $c\in [0,1]$ denote the maximum absolute correlation between the values of $\vg$, i.e. $c \geq |\Sigma_{tt^\prime}| / \sqrt{\Sigma_{tt} \Sigma_{t^\prime t^\prime}} ~~ \forall t\ne t^\prime$. Then, we show that the mean squared error of the hypergradients with respect to $\vmu$ when sharing $W$ contiguous hyperparameters has an upper bound:
\begin{align}
\mathrm{MSE}_W &\leq \frac{(1+c(W-1))}{W}\mathrm{MSE}_1 +\epsilon^2\frac{(W^2-1)}{12} \\
\text{where}~~~~~ \mathrm{MSE}_1 &= \frac{1}{T} \sum_t \mSigma_{tt}
\end{align}
and so for sufficiently small $\epsilon$ and $c$ we have with certainty, $MSE_W<MSE_1$ for some $W>1$, where $MSE_1$ is the hypergradient error without any sharing (See Appendix B for a proof).
\label{theorem}
\end{theorem}

\paragraph{Discussion} Theorem \ref{theorem} demonstrates how sharing $W$ contiguous hyperparameters has two effects: 1) it reduces the hypergradient noise by a factor $\mathcal{O}(W/(1+cW))$ due to the averaging of noisy hypergradients, and 2) it increases the error up to an amount $\mathcal{O}(\epsilon^2W^2)$ due to an induced bias. Intuitively, averaging contiguous hypergradients maximally reduces noise when they aren't correlated ($c=0$), and minimally increases bias when they are drawn from distributions of similar means ($\epsilon=0$). In the simpler case where hypergradients are iid and have the same variance at each step, namely $\vg\sim \mathcal{N}(\vmu,\sigma^2\vone)$, the expressions above become $\mathrm{MSE}_1 = \sigma^2$ and $\mathrm{MSE}_W \leq \mathrm{MSE}_1/W +\epsilon^2(W^2-1)/12$. Note that in all cases, the upper bound on $\mathrm{MSE}_W$ has a single minimum $W^*$ corresponding to the optimal trade-off between noise reduction and bias increase.

\subsection{The FDS algorithm}

As it is presented in Section \ref{sec: forward mode diff background}, forward-mode differentiation would still have a memory cost that scales as $\mathcal{O}(DT)$ since we are learning one hyperparameter per step. However, addressing gradient degradation by sharing hyperparameters also conveniently reduces that memory cost by a factor $W$ down to $\mathcal{O}(DN)$, where $N = T/W$ is the number of unique hyperparameter values we learn. This is because we can safely average hypergradients without calculating them individually, by reusing the same column of $\mZ$ for $W$ contiguous steps. This is shown in Algorithm 1, when learning a schedule of $N^{\bo{\alpha}}$ learning rates with FDS. Here, $\mZ^{\bo{\alpha}}_{[i]}$ refers to the i-th column of matrix $\mZ^{\bo{\alpha}}$. We don't need to store $\bo{\mathcal{H}}$ or $\mA^{\bo{\alpha}}$ in memory since we calculate the Hessian matrix product $\bo{\mathcal{H}}\mZ^{\bo{\alpha}}$ directly. Most importantly, note that hyperparameters aren't updated greedily or online, but are updated once per outer step, which corresponds to differentiating through the entire unrolled inner loop optimization and getting the exact hypergradients $\partial \mathcal{L}_{\text{val}}(\vtheta_T) / \partial \bo{\alpha}$. 

\begin{wrapfigure}[25]{R}{0.5\textwidth}
\begin{minipage}[b]{0.5\textwidth}
\vspace{-0.85cm}
\begin{algorithm}[H]

     \caption{Simplified FDS algorithm when learning $N^\alpha$ learning rates for the SGD optimizer with momentum. Each learning rate is shared over $W$ contiguous time steps}
     \label{alg:forward_mode}

\begin{algorithmic}
\STATE {\bfseries Initialize:} $ N^\alpha, W = T/N^\alpha,\bo{\alpha} = \bo{0}^{N^\alpha}$
\vspace{.20cm}

\STATE {\color{CadetBlue} \textit{\#outer loop}}
 \FOR{$o$ in $1, 2, ...$} 
        \STATE {\bfseries Initialize:} $\mathcal{D}_{\text{train}}$, $\mathcal{D}_{\text{val}}$, $\bo{\theta}_0 \in \mathbb{R}^D$,  \\$\mZ^{\bo{\alpha}} = \bo{0}^{D \times N^\alpha}$, $C^{\bo{\alpha}} = \bo{0}^{D \times N^\alpha}$

        \vspace{.20cm}
        \STATE {\color{CadetBlue}\textit{\#inner loop}}
        \FOR{$t$ in $1, 2, ..., T$}
              \STATE $\bo{x}_{\text{train}}, \bo{y}_{\text{train}} \sim \mathcal{D}_{\text{train}}$
              \STATE $\bo{g}_{\text{train}} = \partial\mathcal{L}_{\text{train}}(\bo{x}_{\text{train}}, \bo{y}_{\text{train}}) / \partial \bo{\theta}$
              
              \vspace{.20cm}
              \STATE {\color{CadetBlue} \textit{\#hyperparameter sharing}}
              \STATE $i = \ceil{t / W}$
              \STATE $\bo{\mathcal{H}}\mZ^{\bo{\alpha}}_{[1:i]} = \partial(\bo{g}_{\text{train}}  \mZ^{\bo{\alpha}}_{[1:i]}) / \partial \bo{\theta}$
              \STATE $\mZ^{\bo{\alpha}}_{[1:i]} = \mA^{\bo{\alpha}}\mZ^{\bo{\alpha}}_{[1:i]} + \mB^{\bo{\alpha}}_{[1:i]}$
              
              \vspace{.20cm}
              \STATE update $C^{\bo{\alpha}}$  (Eq \ref{eq:recursive_alpha})
              \STATE $\bo{\theta}_{t+1} = \Phi(\bo{\theta}_{t}, \bo{g}_{\text{train}})$ 
    \vspace{.20cm}
    \ENDFOR
    
    \vspace{.20cm}
    \STATE $\bo{g}_{\text{val}} = \partial\mathcal{L}_{\text{val}}(\mathcal{D}_{\text{val}}) / \partial \bo{\theta}$
    \STATE $\bo{\alpha} \leftarrow \bo{\alpha} - 0.1 \times \bo{g}_{\text{val}} \mZ^{\bo{\alpha}}/W$
\vspace{.20cm}
\ENDFOR

\end{algorithmic}
\end{algorithm}
\end{minipage}
\end{wrapfigure}

The main cost of Algorithm 1 comes from calculating $\bo{\mathcal{H}}\mZ^{\bo{\alpha}}$. There exists several methods to approximate this product, but we found them too crude for long horizons. This includes first-order approximations or truncation \cite{Shaban2018TruncatedBackPropBilevelOpt}, which can be adapted to FDS trivially. One could also use functional forms for more complex schedules to be learned in terms of fewer hyperparameters, but this typically makes stronger assumptions about the shape of each hyperparameter schedule, which can easily cloud the true performance of HPO algorithms. In practice, we calculate $\bo{\mathcal{H}}\mZ^{\bo{\alpha}}$ exactly, and use a similar form to Algorithm 1 to learn $\bo{\alpha}$, $\bo{\beta}$ and $\bo{\xi}$.

\begin{figure*}[t!]
    \centering
    \includegraphics[width=\textwidth]{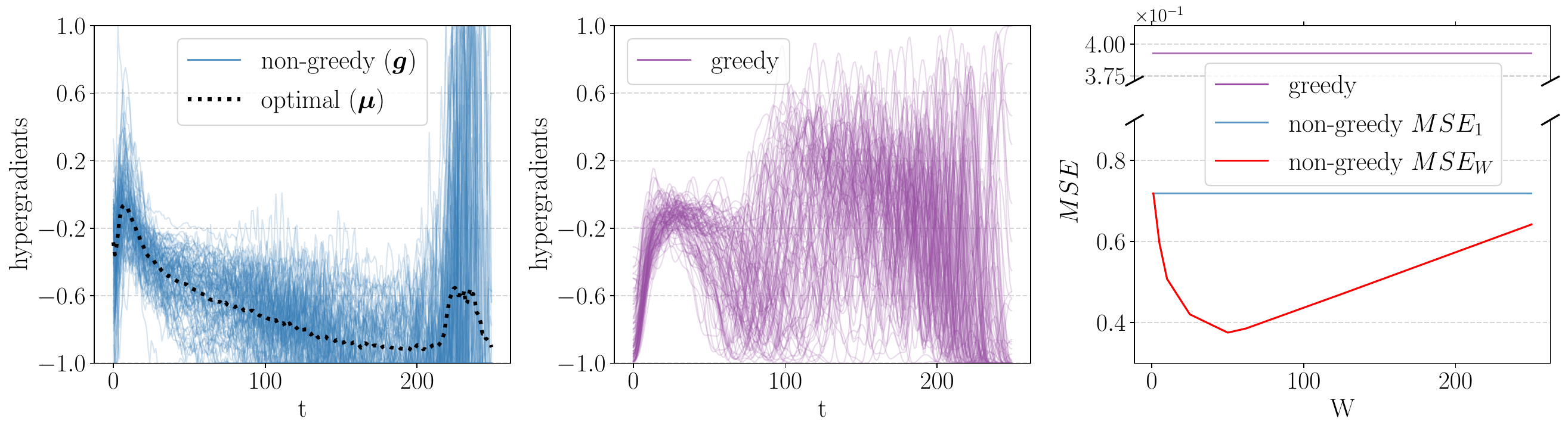}
    \caption{ Hypergradients of $\bo{\alpha}$ on SVHN for 100 seeds in the non-greedy (left) and greedy (middle) setting. The mean squared error is also shown (right), and is calculated with respect to the optimal hypergradient ($\vmu$), i.e. the ensemble average of non-greedy hypergradients (dotted line in left figure). We can see that sharing hyperparameters over $W$ steps lowers the $MSE$ even for the small value of $T=250$ used here. The best trade-off between noise reduction and bias increase is $W=50$. }
    \label{fig:hypergrads_and_MSE}
\end{figure*}
%TODO put fire with larger slope for bigger W
\section{Experiments}

Our experiments show how FDS mitigates gradient degradation and outperforms competing HPO methods for tasks with a long horizon. In Sections \ref{sec:exp_MSE_reduction} and \ref{sec:MNIST and SVHN experiments} we consider small datasets (MNIST and SVHN) and a small network (LeNet) to make reverse-mode differentiation tractable, so that the effect of hyperparameter sharing can be directly measured. In Sections \ref{sec:exp_cifar} and \ref{sec:exp_HPO_baselines}, we then showcase FDS on CIFAR-10 where only forward-mode differentiation is tractable. All experiments are carried out on a single GTX 2080 GPU. More implementation details can be found in Appendix C.

\subsection{The effect of hyperparameter sharing on hypergradient noise}
\label{sec:exp_MSE_reduction}

We consider a LeNet network trained on an inner loop of $T=250$ gradient steps, and calculate the hypergradients of the learning rate $\bo{\alpha}$ across $100$ seeds. Each seed is evaluated at the same value of $\bo{\alpha}$, but corresponds to a different training dataset ordering and weight initialization. We can calculate the hypergradients greedily ($H=1$) and non-greedily ($H=T$). The optimal hypergradients are considered to be the ensemble average of the non-greedy seeds as per Section \ref{subsec:sharing_hyperparameters}, which allows an MSE to be calculated for each method. These results are shown in Figure \ref{fig:hypergrads_and_MSE} for a learning rate schedule initialized to small values. We observe that greediness is a poor approximation to the optimal hypergradients, and that time averaging contiguous hypergradients in the non-greedy case can significantly reduce the MSE even when averaging over only $W=50$ steps. 

The simplicity of this problem allows us to calculate $\mSigma$, $c$ and $\epsilon$ as defined in Theorem \ref{theorem}, to verify that our upper bound holds and that its shape as a function of $W$ is realistic. In the setting shown in Figure \ref{fig:hypergrads_and_MSE} we find $\epsilon = \max |\mu_{t+1} - \mu_t|\sim 0.08$ and $(1/T) \sum_t \mSigma_{tt} \sim 0.25$. The value of the maximum correlation between any two steps, $c$, can be quite high which makes the upper bound loose. In practice however, the shape of the upper bound in Theorem \ref{theorem} as a function of $W$ (as plotted in Appendix B) closely matches that of the measured MSE shown in Figure \ref{fig:hypergrads_and_MSE}. Note that the values of $\epsilon$, $\mSigma,$ and $c$ can vary depending on the value of $\bo{\alpha}$. As illustrated in Theorem \ref{theorem}, we find that settings that have a smaller values of $\epsilon_t$ benefit from a larger $W$ and reduce the MSE more (see Appendix D for more examples).

\begin{figure*}[t!]
    \centering
    \includegraphics[width=\textwidth]{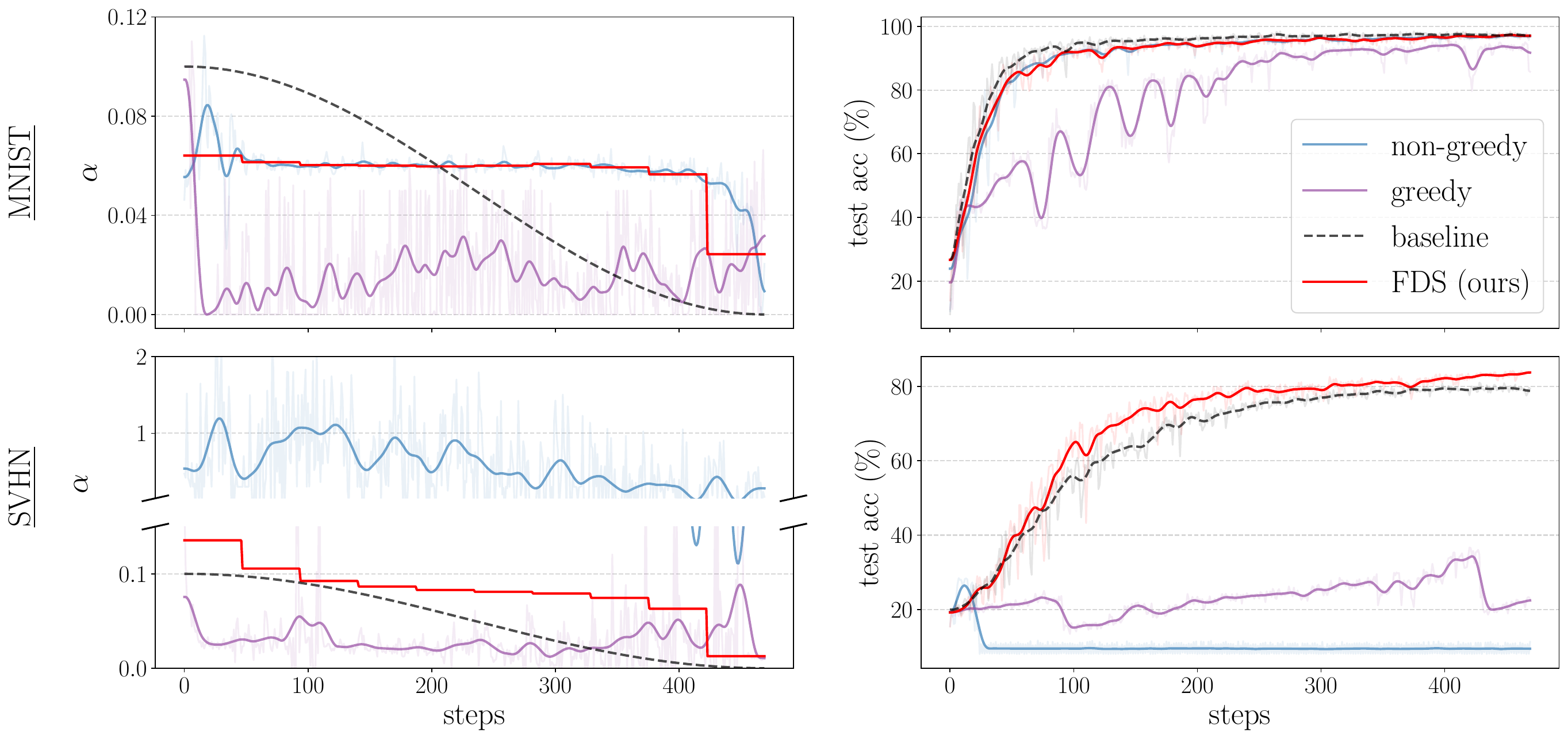}
    \caption{The learning rate schedule $\bo{\alpha}$ learned for the MNIST and SVHN datasets using a LeNet architecture over one epoch. We observe that on real-world datasets like SVHN, both greedy and non-greedy hyperparameter optimizations fail to learn decent learning rate schedules when using one hypergradient per inner step. However, sharing learning rates over contiguous steps stabilizes non-greedy hypergradients and allows us to find learning rates that can even outperform reasonable off-the-shelf schedules in this setting.}
    \label{fig:reverse_lrs}
\end{figure*}

\subsection{The effect of hyperparameter sharing on HPO}
\label{sec:MNIST and SVHN experiments}

In the section above, we considered hypergradient noise around a fixed hyperparameter setting. In this section, we consider how that noise and its mitigation translate to meta-learning hyperparameters over several outer steps. 

In Figure \ref{fig:reverse_lrs}, we initialize $\bo{\alpha} = \bo{0}$ and train a LeNet network over 1 epoch ($T\sim 500$ gradient steps) on MNIST and SVHN. We compare the maximally greedy setting ($H=1$), the non-greedy setting ($H=T$), and FDS (also $H=T$ but with sharing of $W \sim 50$ contiguous hyperparameters). In all cases we take $50$ outer steps per hyperparameter. We make greediness more transparent by not using tricks as in Hypergradient Descent \cite{Baydin2018HypergradientDescent}, where $\alpha_t$ is set to $\alpha_{t-1}$ before its hypergradients are calculated. As previously observed by \cite{Wu2018ShortHorizonBias}, greedy optimization leads to poor solutions with learning rates that are always too small. While the non-greedy setting without sharing works well for simple datasets like MNIST, it fails for real-world datasets like SVHN, converging to much higher learning rates than reasonable. This is due to gradient degradation, whose negative effect can compound during outer optimization, as a single large learning rate can increase the hypergradient noise for neighbouring steps. As we share hyperparameters in FDS, we reduce and stabilize the outer optimization. This allows us to learn a much more sensible learning rate schedule, which can even outperform a reasonable cosine annealing off-the-shelf schedule on SVHN.

\begin{figure}[t] %!htbp
\centering
\includegraphics[width=\textwidth]{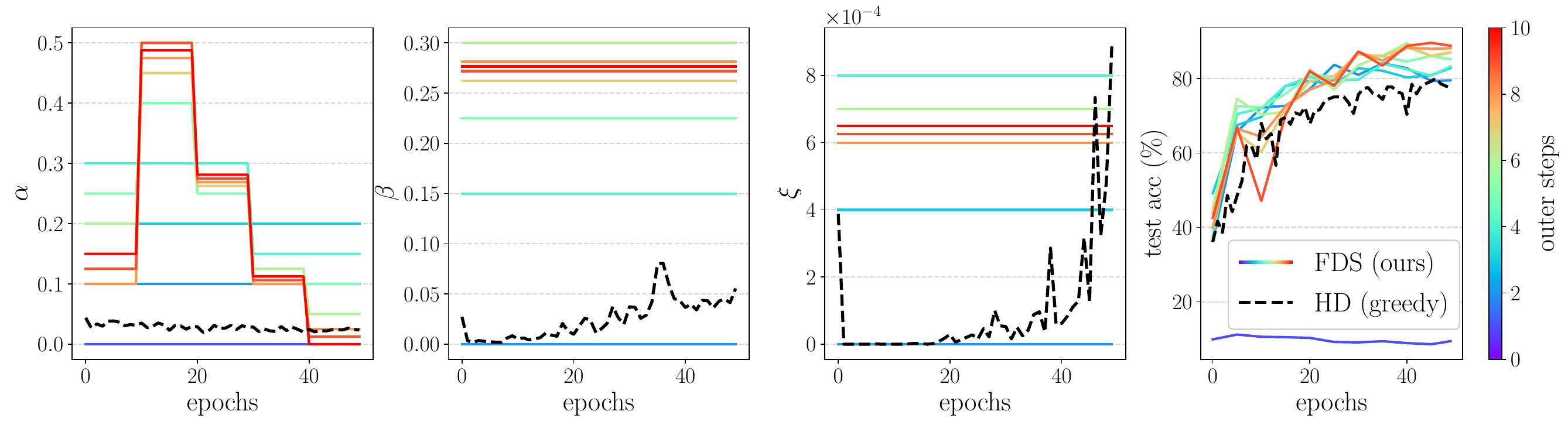}
\caption{FDS applied to the SGD optimizer to learn (from left to right) the learning rate schedule $\bo{\alpha}$, the momentum $\bo{\beta}$, and weight decay $\bo{\xi}$ for a WRN-16-1 on CIFAR-10. For each outer step (color) we solve CIFAR-10 from scratch for $50$ epochs, and update all hyperparameters such that the final weights minimize some validation loss. We use hyperparameter sharing over $W = 10$, $50$ and $50$ epochs for $\bo{\alpha}$, $\bo{\beta}$ and $\bo{\xi}$ respectively. All hyperparameters are initialized to zero and converge within just $10$ outer steps to values that significantly outperform Hypergradient Descent (HD) \cite{Baydin2018HypergradientDescent}, the greedy alternative. We match the performance of the best known hyperparameters in that setting and do so much faster than state-of-the-art black-box methods (see Section \ref{sec:exp_HPO_baselines}).}
\label{fig:forward_lrs_cifar10}
\end{figure}

\subsection{FDS on CIFAR-10}
\label{sec:exp_cifar}

We demonstrate that our algorithm can be used to learn the learning rate, momentum and weight decay over $50$ epochs of CIFAR-10 ($H \sim 10^4$) for a WideResNet of 16 layers (WRN-16-1). We choose not to use larger architectures or more epochs to enable compute time for more extensive comparisons and because hyperparameters matter most for fewer epochs. Note also that we are not interested in finding the architecture with the best performance, but rather in finding the best hyperparameters given an architecture. For the learning rate, we choose $W$ such that the ratio of $T/W$ is similar to the optimal one found in \ref{sec:exp_MSE_reduction}, and we set $W=T$ for the momentum and weight decay since only a single value is commonly used for these hyperparameters. The schedules learned are shown in Figure \ref{fig:forward_lrs_cifar10}, which demonstrates that FDS converges in just $10$ outer steps to hyperparameters that are very different to online greedy differentiation \cite{Baydin2018HypergradientDescent}, and correspond to significantly better test accuracy performances. Note that having the maximum learning rate be large and occur half way during training is reminiscent of the one-cycle schedule \cite{Leslie2017SuperConvergence}. 

\subsection{FDS compared to other HPO methods}
\label{sec:exp_HPO_baselines}

A common theme in meta-learning research has been the lack of appropriate baselines, with researchers often finding that random search (RS) can outperform complex search algorithms, for instance in NAS \cite{Li2019RSforNAS} or automatic augmentation \cite{Cubuk2020RandaugmentPA}. In this section we compare FDS to several competing HPO methods on CIFAR-10, in both the performance of the hyperparameters found and the time it takes to find them. We consider Random Search (RS), Bayesian Optimization (BO), Hypergradient Descent (HD) \cite{Baydin2018HypergradientDescent}, HyperBand (HB) \cite{Li2017Hyperband} and the combination of BO and HB, namely BOHB \cite{Falkner2018BOHB}. The latter is typically regarded as state-of-the-art in HPO. We use the official HpBandster \cite{HpBandster} implementation of these algorithms, except for HD which we re-implemented ourselves. 

The performance of HPO methods is often clouded by very small search ranges. For instance, in \cite{Falkner2018BOHB} the learning rate is searched over the range $[10^{-6}, 10^{-1}]$. In the case of DARTS \cite{Liu2019DARTS}, expanding the search space to include many poor architectures has helped diagnose issues with its search algorithm \cite{Zela2020RobustifyingDARTS}. For these reasons, we assume only weak priors and consider large search ranges: $\alpha \in [-1,1]$, $\beta \in [-1.5, 1.5]$, and $\xi \in [-4\times10^{-3}, 4\times10^{-3}]$, which includes many poor hyperparameter values. In FDS we can specify search ranges by using a fixed update size $\bo{\gamma}$ to the hyperparameters, which decays by $2$ every time the hypergradient flips sign, which is common in sign based optimizers \cite{Jacobs1988DecayLearningRateFromSign, Riedmiller1993RPROP, Bernstein2018SignSGD, Safaryan2019StochasticSignDescentMethods}. We found that black-box HPO methods did not scale well to more than $\sim 10 $ hyperparameters for such large search ranges, and so considered learning $7$ learning rates, $1$ momentum and $1$ weight decay over $50$ epochs. Note that increasing these numbers doesn't affect the accuracy of FDS but significantly reduces that of RS, BO, HB and BOHB. The performance over time of the hyperparameters found by each method is shown in Figure \ref{fig:perf over time}. As is common in HPO, we plot regret (in test accuracy) with respect to the best hyperparameters known in this setting, which we obtained from an expensive grid search around the most common hyperparameters used in the literature (more details in Appendix F). To squeeze the optimal performance out of FDS in this experiment, we match the process used in HB and BOHB, namely using smaller budgets for some runs, in particular early ones. We can see that our method reaches zero regret in just $3$ hours, while the next best method (BOHB) reaches zero regret in $60$ hours. Other methods did not reach zero regret in the maximum of $100$ hours they were run for. Note also that we retain the convergence speed of online greedy differentiation \cite{Baydin2018HypergradientDescent} while outperforming its regret by $\sim 10\%$ test accuracy.

%------------------------------------------
%\begin{wraptable}{r}{8.5cm} %tables must be centered
% \begin{wraptable}{r}{0.5\textwidth}

% \caption{Final test accuracy of the hyperparameters found with several HPO methods, as well as the time to find them. FDS outperforms other methods in both speed and accuracy. Errors shown are standard deviations over $8$ seeds.}
% \label{tab:baselines}
% \begin{center}
% \begin{tabular}{l|ccc}
% Method & \thead{Convergence \\ time (hours)} &  \thead{Test acc \\ ($\%$)}
% \\\midrule

% RS   & $> 100$ & $80.8$ \tiny{$\mypm 4.9$} \\

% BO & $> 100$ & $84.5$ \tiny{$\mypm 5.2$} \\

% HB \cite{Li2017Hyperband} & 65 & $84.1$ \tiny{$\mypm 4.1$} \\

% BOHB \cite{Falkner2018BOHB}  & 55 & $89.1$ \tiny{$\mypm 0.5$}  \\

% HD \cite{Baydin2018HypergradientDescent} & 5 & $80.6$ \tiny{$\mypm 1.0$} \\

% FDS (Ours) & $\bo{3}$ & $\bo{89.4}$ \tiny{$\mypm 0.3$} \\

% %1\bottomrule
% \end{tabular}
% \end{center}

% %\end{wraptable}

% \end{wraptable}

\section{Discussion}

While FDS is significantly better than modern HPO alternatives in the case of learning differentiable hyperparameters over long horizons, it is worth pointing out some of its limitations. First, black-box methods are slower but can readily tackle non-differentiable hyperparameters, while FDS would need to use relaxation techniques to differentiate through, say, discrete variables. However, forward-mode differentiation can provide exact hypergradients for any differentiable hyperparameter, contrary to some other approaches like implicit differentiation which approximates hypergradients, and does so for some types of differentiable hyperparameters only. Then, the computational and memory cost of FDS scales linearly with the number of hyperparameters. For instance, for a WideResNet of $16$ layers we are limited in memory to $\sim 10^3$ hyperparameters on a single $12$ GB GPU, which falls short of reverse-mode-based greedy methods which scale to millions of hyperparameters. In summary, the main advantage of FDS over black-box methods is its convergence speed, and its main advantage over reverse-mode methods is that its memory cost is constant with horizon size, which enables differentiating non-greedily. 

 While \textit{automatic} forward-mode differentiation is becoming available for some deep learning libraries like JAX \cite{Bradbury2018JAX}, it remains unavailable for the most popular libraries which means that some pen-and-paper work is still required to derive hypergradients for given hyperparameters (Eq \ref{eq:recursive_alpha}). The focus of our future work will be on the many meta-learning applications that rely on greedy optimization, such as differentiable architecture search, to improve their performance by making them non-greedy.

\section{Conclusion}

This work makes an important step towards gradient-based HPO for long horizons by introducing FDS, which enables non-greediness through the mitigation of gradient degradation. More specifically, we theorize and demonstrate that sharing hyperparameters over contiguous time steps is a simple yet efficient way to reduce the error in their hypergradients; a setting which naturally lands itself to forward-mode differentiation. Finally, we show that FDS outperform greedy gradient-based alternatives in the quality of hyperparameters found, while being significantly faster than all state-of-the-art black-box methods. We hope that our work encourages the community to reconsider gradient-based HPO in terms of non-greediness, and pave the way towards a universal hyperparameter solver.

\clearpage
\paragraph{Acknowledgements}

The authors would like to thank Joseph Mellor, Antreas Antoniou, Miguel Jaques, Luke Darlow and Benjamin Rhodes for their useful feedback throughout this project. Our work was supported in part by the EPSRC Centre for Doctoral Training in Data Science, funded by the UK Engineering and Physical Sciences Research Council (grant EP/L016427/1) and the University of Edinburgh, as well as a Huawei DDMPLab Innovation Research Grant.

\bibliography{bibliography}

\begin{thebibliography}{10}

\bibitem{He2015ResNet}
K.~He, X.~Zhang, S.~Ren, and J.~Sun, ``Deep residual learning for image
  recognition,'' in {\em Proceedings of the IEEE conference on computer vision
  and pattern recognition}, 2016.

\bibitem{Bock2018BigGan}
A.~Brock, J.~Donahue, and K.~Simonyan, ``Large scale {GAN} training for high
  fidelity natural image synthesis,'' in {\em International Conference on
  Learning Representations}, 2019.

\bibitem{Devlin2018BERT}
J.~Devlin, M.-W. Chang, K.~Lee, and K.~Toutanova, ``Bert: Pre-training of deep
  bidirectional transformers for language understanding,'' {\em arXiv preprint
  arXiv:1810.04805}, 2018.

\bibitem{VanDenOord2018WaveNet}
A.~v.~d. Oord, S.~Dieleman, H.~Zen, K.~Simonyan, O.~Vinyals, A.~Graves,
  N.~Kalchbrenner, A.~Senior, and K.~Kavukcuoglu, ``Wavenet: A generative model
  for raw audio,'' {\em arXiv preprint arXiv:1609.03499}, 2016.

\bibitem{Kingma2015Adam}
D.~P. Kingma and J.~Ba, ``Adam: {A} method for stochastic optimization,'' in
  {\em International Conference on Learning Representations} (Y.~Bengio and
  Y.~LeCun, eds.), 2015.

\bibitem{Loshchilov2017WarmRestarts}
I.~Loshchilov and F.~Hutter, ``Sgdr: Stochastic gradient descent with warm
  restarts,'' in {\em International Conference on Learning Representations},
  2017.

\bibitem{Li2020BudgetedSchedules}
M.~Li, E.~Yumer, and D.~Ramanan, ``Budgeted training: Rethinking deep neural
  network training under resource constraints,'' in {\em International
  Conference on Learning Representations}, 2020.

\bibitem{Goyal2017LargeMinibatchSGD}
P.~Goyal, P.~Doll{\'a}r, R.~Girshick, P.~Noordhuis, L.~Wesolowski, A.~Kyrola,
  A.~Tulloch, Y.~Jia, and K.~He, ``Accurate, large minibatch sgd: Training
  imagenet in 1 hour,'' {\em arXiv preprint arXiv:1706.02677}, 2017.

\bibitem{Dong2020AutoHAS}
X.~Dong, M.~Tan, A.~W. Yu, D.~Peng, B.~Gabrys, and Q.~V. Le, ``Autohas:
  Differentiable hyper-parameter and architecture search,'' 2020.

\bibitem{loshchilov2018AdamW}
I.~Loshchilov and F.~Hutter, ``Decoupled weight decay regularization,'' in {\em
  International Conference on Learning Representations}, 2019.

\bibitem{Ioffe2015BatchNorm}
S.~Ioffe and C.~Szegedy, ``Batch normalization: Accelerating deep network
  training by reducing internal covariate shift,'' in {\em International
  conference on Machine Learning} (F.~Bach and D.~Blei, eds.), Proceedings of
  Machine Learning Research, (Lille, France), PMLR, 07--09 Jul 2015.

\bibitem{Li2017Hyperband}
L.~Li, K.~G. Jamieson, G.~DeSalvo, A.~Rostamizadeh, and A.~Talwalkar,
  ``Hyperband: {A} novel bandit-based approach to hyperparameter
  optimization,'' {\em J. Mach. Learn. Res.}, vol.~18, pp.~185:1--185:52, 2017.

\bibitem{Falkner2018BOHB}
S.~Falkner, A.~Klein, and F.~Hutter, ``{BOHB}: Robust and efficient
  hyperparameter optimization at scale,'' in {\em Proceedings of the 35th
  International Conference on Machine Learning}, pp.~1436--1445, 2018.

\bibitem{Baydin2018HypergradientDescent}
A.~G. Baydin, R.~Cornish, D.~M. Rubio, M.~Schmidt, and F.~Wood, ``Online
  learning rate adaptation with hypergradient descent,'' in {\em International
  Conference on Learning Representations}, 2018.

\bibitem{Wu2018ShortHorizonBias}
Y.~Wu, M.~Ren, R.~Liao, and R.~Grosse., ``Understanding short-horizon bias in
  stochastic meta-optimization,'' in {\em International Conference on Learning
  Representations}, 2018.

\bibitem{Franceschi2017ForwardAndReverseGradientBasedHO}
L.~Franceschi, M.~Donini, P.~Frasconi, and M.~Pontil, ``Forward and reverse
  gradient-based hyperparameter optimization,'' in {\em International
  conference on Machine Learning} (D.~Precup and Y.~W. Teh, eds.), Proceedings
  of Machine Learning Research, (International Convention Centre, Sydney,
  Australia), PMLR, 06--11 Aug 2017.

\bibitem{Donini2019SchedulingLearningRateNewInsights}
M.~{Donini}, L.~{Franceschi}, M.~{Pontil}, O.~{Majumder}, and P.~{Frasconi},
  ``{Scheduling the Learning Rate via Hypergradients: New Insights and a New
  Algorithm},'' {\em arXiv e-prints}, Oct. 2019.

\bibitem{Feurer2019HyperparamOptBookChapter}
M.~Feurer and F.~Hutter, {\em Chapter 1: Hyperparameter Optimization}.
\newblock Cham: Springer International Publishing, 2019.

\bibitem{Snoek2015ScalableBayesianOptimizationDeepNets}
J.~Snoek, O.~Rippel, K.~Swersky, R.~Kiros, N.~Satish, N.~Sundaram, M.~Patwary,
  M.~Prabhat, and R.~Adams, ``Scalable bayesian optimization using deep neural
  networks,'' in {\em International conference on Machine Learning}, 2015.

\bibitem{Zoph2017NASwithRL}
B.~Zoph and Q.~V. Le, ``Neural architecture search with reinforcement
  learning,'' in {\em International Conference on Learning Representations},
  2017.

\bibitem{Jaderberg2017PopBasedTrainingNets}
M.~Jaderberg, V.~Dalibard, S.~Osindero, W.~M. Czarnecki, J.~Donahue, A.~Razavi,
  O.~Vinyals, T.~Green, I.~Dunning, K.~Simonyan, {\em et~al.}, ``Population
  based training of neural networks,'' {\em arXiv preprint arXiv:1711.09846},
  2017.

\bibitem{Bengio2000GradientBasedHyperparamOpt}
Y.~Bengio, ``Gradient-based optimization of hyperparameters,'' {\em Neural
  Comput.}, Aug. 2000.

\bibitem{Hospedales2020MetaLearningSurvey}
T.~Hospedales, A.~Antoniou, P.~Micaelli, and A.~Storkey, ``Meta-learning in
  neural networks: A survey,'' {\em arXiv preprint arXiv:2004.05439}, 2020.

\bibitem{Finn2017MAML}
C.~Finn, P.~Abbeel, and S.~Levine, ``Model-agnostic meta-learning for fast
  adaptation of deep networks,'' in {\em Proceedings of the 34th International
  Conference on Machine Learning, {ICML} 2017, Sydney, NSW, Australia, 6-11
  August 2017} (D.~Precup and Y.~W. Teh, eds.), vol.~70 of {\em Proceedings of
  Machine Learning Research}, pp.~1126--1135, {PMLR}, 2017.

\bibitem{Franceschi2018BilevelPF}
L.~Franceschi, P.~Frasconi, S.~Salzo, R.~Grazzi, and M.~Pontil, ``Bilevel
  programming for hyperparameter optimization and meta-learning,'' in {\em
  International conference on Machine Learning}, 2018.

\bibitem{Werbos1990BackpropagationThroughTime}
P.~J. {Werbos}, ``Backpropagation through time: what it does and how to do
  it,'' {\em Proceedings of the IEEE}, 1990.

\bibitem{Williams1989AlgorithmRecurrentNeuralNetwork}
R.~J. {Williams} and D.~{Zipser}, ``A learning algorithm for continually
  running fully recurrent neural networks,'' {\em Neural Computation}, 1989.

\bibitem{Domke2012OptimizationBasedModeling}
J.~Domke, ``Generic methods for optimization-based modeling,'' in {\em
  International Conference on Artificial Intelligence and Statistics} (N.~D.
  Lawrence and M.~Girolami, eds.), Proceedings of Machine Learning Research,
  (La Palma, Canary Islands), PMLR, 21--23 Apr 2012.

\bibitem{Maclaurin2015GradientHO}
D.~{Maclaurin}, D.~{Duvenaud}, and R.~P. {Adams}, ``{Gradient-based
  Hyperparameter Optimization through Reversible Learning},'' {\em arXiv
  e-prints}, Feb. 2015.

\bibitem{Pedregosa2016GradientHO}
F.~Pedregosa, ``Hyperparameter optimization with approximate gradient,'' in
  {\em International conference on Machine Learning}, 2016.

\bibitem{Fu2016DrMAD}
J.~Fu, H.~Luo, J.~Feng, K.~H. Low, and T.-S. Chua, ``Drmad: Distilling
  reverse-mode automatic differentiation for optimizing hyperparameters of deep
  neural networks,'' in {\em IJCAI}, 2016.

\bibitem{Shaban2018TruncatedBackPropBilevelOpt}
A.~Shaban, C.-A. Cheng, N.~Hatch, and B.~Boots, ``Truncated back-propagation
  for bilevel optimization,'' in {\em AISTATS}, 2019.

\bibitem{Larsen1996OptimalUseOfValidation}
J.~{Larsen}, L.~K. {Hansen}, C.~{Svarer}, and M.~{Ohlsson}, ``Design and
  regularization of neural networks: the optimal use of a validation set,'' in
  {\em Neural Networks for Signal Processing VI. Proceedings of the 1996 IEEE
  Signal Processing Society Workshop}, pp.~62--71, 1996.

\bibitem{Rajeswaran2019ImplicitMAML}
A.~Rajeswaran, C.~Finn, S.~M. Kakade, and S.~Levine, ``Meta-learning with
  implicit gradients,'' in {\em Advances in Neural Information Processing
  Systems}, Curran Associates, Inc., 2019.

\bibitem{Lorraine2019MillionsHyperparamOptimImplicit}
J.~{Lorraine}, P.~{Vicol}, and D.~{Duvenaud}, ``{Optimizing Millions of
  Hyperparameters by Implicit Differentiation},'' {\em arXiv e-prints}, Nov.
  2019.

\bibitem{Metz2019PathologiesInTrainingOptimizers}
L.~Metz, N.~Maheswaranathan, J.~Nixon, D.~Freeman, and J.~Sohl-Dickstein,
  ``Understanding and correcting pathologies in the training of learned
  optimizers,'' vol.~97, pp.~4556--4565, 09--15 Jun 2019.

\bibitem{Luketina2016GradBasedRegLearning}
J.~Luketina, M.~Berglund, K.~Greff, and T.~Raiko, ``Scalable gradient-based
  tuning of continuous regularization hyperparameters,'' in {\em International
  conference on Machine Learning}, 2016.

\bibitem{Liu2019DARTS}
H.~Liu, K.~Simonyan, and Y.~Yang, ``{DARTS}: Differentiable architecture
  search,'' in {\em International Conference on Learning Representations},
  2019.

\bibitem{Wang2018DatasetDistillation}
T.~Wang, J.~Zhu, A.~Torralba, and A.~A. Efros, ``Dataset distillation,'' {\em
  CoRR}, vol.~abs/1811.10959, 2018.

\bibitem{Ren2018LearningToReweight}
M.~Ren, W.~Zeng, B.~Yang, and R.~Urtasun, ``Learning to reweight examples for
  robust deep learning,'' in {\em International conference on Machine
  Learning}, 2018.

\bibitem{Bengio1993RNNGradientIssues}
Y.~{Bengio}, P.~{Frasconi}, and P.~{Simard}, ``The problem of learning
  long-term dependencies in recurrent networks,'' in {\em IEEE International
  Conference on Neural Networks}, 1993.

\bibitem{Bengio1994LongTermGradInefficient}
Y.~{Bengio}, P.~{Simard}, and P.~{Frasconi}, ``Learning long-term dependencies
  with gradient descent is difficult,'' {\em IEEE Transactions on Neural
  Networks}, 1994.

\bibitem{Parmas2018PIPPSCurseOfChaos}
P.~Parmas, C.~E. Rasmussen, J.~Peters, and K.~Doya, ``{PIPPS}: Flexible
  model-based policy search robust to the curse of chaos,'' vol.~80 of {\em
  Proceedings of Machine Learning Research}, (Stockholmsmässan, Stockholm
  Sweden), pp.~4065--4074, PMLR, 10--15 Jul 2018.

\bibitem{Hochreiter1997LSTM}
S.~Hochreiter and J.~Schmidhuber, ``Long short-term memory,'' {\em Neural
  computation}, 1997.

\bibitem{Pascanu2013DifficultyOfRNNs}
R.~Pascanu, T.~Mikolov, and Y.~Bengio, ``On the difficulty of training
  recurrent neural networks,'' in {\em International conference on Machine
  Learning}, 2013.

\bibitem{Flennerhag2020WarpedGradientDescent}
S.~Flennerhag, A.~A. Rusu, R.~Pascanu, F.~Visin, H.~Yin, and R.~Hadsell,
  ``Meta-learning with warped gradient descent,'' in {\em International
  Conference on Learning Representations}, 2020.

\bibitem{Zagoruyko2016WRN}
S.~Zagoruyko and N.~Komodakis, ``Wide residual networks,'' in {\em BMVC}, 2016.

\bibitem{Stackelberg1952BilevelOptimization}
H.~Stackelberg, {\em {The Theory Of Market Economy}}.
\newblock Oxford University Press, 1952.

\bibitem{Paszke2019Pytorch}
A.~Paszke, S.~Gross, F.~Massa, A.~Lerer, J.~Bradbury, G.~Chanan, T.~Killeen,
  Z.~Lin, N.~Gimelshein, L.~Antiga, A.~Desmaison, A.~Kopf, E.~Yang, Z.~DeVito,
  M.~Raison, A.~Tejani, S.~Chilamkurthy, B.~Steiner, L.~Fang, J.~Bai, and
  S.~Chintala, ``Pytorch: An imperative style, high-performance deep learning
  library,'' in {\em Advances in Neural Information Processing Systems}, Curran
  Associates, Inc., 2019.

\bibitem{Walters1982ErgodicTheory}
P.~Walters, {\em An Introduction to Ergodic Theory}.
\newblock Graduate texts in mathematics, Springer-Verlag, 1982.

\bibitem{Boltzmann1896Ergodicity}
L.~Boltzmann, {\em Vorlesungen uber Gastheorie}.
\newblock J.A. Barth, 1896.

\bibitem{Peters2019ErgodicityEconomics}
O.~Peters, ``The ergodicity problem in economics,'' {\em Nature Physics},
  vol.~15, pp.~1216--1221, Dec 2019.

\bibitem{Leslie2017SuperConvergence}
L.~N. Smith and N.~Topin, ``Super-convergence: Very fast training of residual
  networks using large learning rates,'' {\em CoRR}, vol.~abs/1708.07120, 2017.

\bibitem{Li2019RSforNAS}
L.~Li and A.~Talwalkar, ``Random search and reproducibility for neural
  architecture search,'' in {\em UAI}, 2019.

\bibitem{Cubuk2020RandaugmentPA}
E.~D. Cubuk, B.~Zoph, J.~Shlens, and Q.~V. Le, ``Randaugment: Practical
  automated data augmentation with a reduced search space,'' {\em 2020 IEEE/CVF
  Conference on Computer Vision and Pattern Recognition Workshops (CVPRW)},
  pp.~3008--3017, 2020.

\bibitem{HpBandster}
A.~Klein. \url{https://automl.github.io/HpBandSter/build/html/quickstart.html},
  2019.

\bibitem{Zela2020RobustifyingDARTS}
A.~Zela, T.~Elsken, T.~Saikia, Y.~Marrakchi, T.~Brox, and F.~Hutter,
  ``Understanding and robustifying differentiable architecture search,'' in
  {\em International Conference on Learning Representations}, 2020.

\bibitem{Jacobs1988DecayLearningRateFromSign}
``Increased rates of convergence through learning rate adaptation,'' {\em
  Neural Networks}, vol.~1, no.~4, pp.~295--307, 1988.

\bibitem{Riedmiller1993RPROP}
M.~{Riedmiller} and H.~{Braun}, ``A direct adaptive method for faster
  backpropagation learning: the rprop algorithm,'' in {\em IEEE International
  Conference on Neural Networks}, pp.~586--591 vol.1, 1993.

\bibitem{Bernstein2018SignSGD}
J.~Bernstein, Y.~Wang, K.~Azizzadenesheli, and A.~Anandkumar, ``{SIGNSGD:}
  compressed optimisation for non-convex problems,'' in {\em Proceedings of the
  35th International Conference on Machine Learning, {ICML} 2018,
  Stockholmsm{\"{a}}ssan, Stockholm, Sweden, July 10-15, 2018} (J.~G. Dy and
  A.~Krause, eds.), vol.~80 of {\em Proceedings of Machine Learning Research},
  pp.~559--568, {PMLR}, 2018.

\bibitem{Safaryan2019StochasticSignDescentMethods}
M.~{Safaryan} and P.~{Richt{\'a}rik}, ``{On Stochastic Sign Descent Methods},''
  {\em arXiv e-prints}, p.~arXiv:1905.12938, May 2019.

\bibitem{Bradbury2018JAX}
J.~Bradbury, R.~Frostig, P.~Hawkins, M.~J. Johnson, C.~Leary, D.~Maclaurin,
  G.~Necula, A.~Paszke, J.~Vander{P}las, S.~Wanderman-{M}ilne, and Q.~Zhang,
  ``{JAX}: composable transformations of {P}ython+{N}um{P}y programs,'' 2018.

\end{thebibliography}
\bibliographystyle{ieeetr}

%%%%%%%%%%%%%%%%%%%%%%%%%%%%%%%%
\clearpage
\section*{\fontsize{13}{17}\selectfont Appendices}
\subsection*{\fontsize{12}{17}\selectfont Appendix A: Forward-mode Hypergradient Derivations}
\vspace{0.5cm}

Recall that we are interested in calculating

\begin{equation*}
 \bo{Z}_t = \bo{A}_t \bo{Z}_{t-1} + \bo{B}_t
\end{equation*}

recursively during the inner loop, where 

\begin{equation*}
\bo{Z}_t = \dfrac{d\bo{\theta}_t}{d\bo{\lambda}}
\hspace{1cm}
\bo{A}_t = \dfrac{\partial\bo{\theta}_t}{\partial\bo{\theta}_{t-1}}\biggr\rvert_{\bo{\lambda}} 
\hspace{1cm}
\bo{B}_t = \dfrac{\partial\bo{\theta}_t}{\partial\bo{\lambda}}\biggr\rvert_{\bo{\theta}_{t-1}}
\end{equation*}

so that we can calculate the hypergradients on the final step using 

\begin{equation*}
\dfrac{d\mathcal{L}_{\text{val}}}{ d\bo{\lambda}} = \dfrac{\partial\mathcal{L}_{\text{val}}}{ \partial\bo{\theta}_T} \bo{Z}_T
\end{equation*}
Each type of hyperparameter needs its own matrix $\bo{Z}_t$, and therefore its own matrices $\bo{A}_t$, and $\bo{B}_t$. Consider first the derivation of these matrices for the learning rate, namely $\bo{\lambda} = \bo{\alpha}$. Recall that the update rule of SGD with momentum and weight decay after substituting the velocity $\bo{v}_t$ in is 

\begin{equation*}
\bo{\theta}_t = \bo{\theta}_{t-1} - \alpha_t \left( \beta_t \bo{v}_{t-1} + \dfrac{\partial \mathcal{L}_{\text{train}}}{\partial \bo{\theta}_{t-1}} + \xi_t \bo{\theta}_{t-1} \right)
\end{equation*}

and therefore it follows directly that

\begin{equation*}
\bo{A}_t^{\bo{\alpha}} = \vone - \alpha_t \left(\dfrac{\partial^2 \mathcal{L}_{\text{train}}}{\partial \bo{\theta}_{t-1}^2} + \xi_t \vone \right)
\end{equation*}

The calculation of $\bo{B}_t^{\bo{\alpha}}$ is a bit more involved in our work because when using momentum, $\bo{v}_{t-1}$ is now itself a function of $\bo{\alpha}$. First we write

\begin{align*}
\bo{B}_t^{\bo{\alpha}} & = -\beta_t \left( \dfrac{\partial \alpha_t}{\partial \bo{\alpha}} \bo{v}_{t-1} + \alpha_t \dfrac{\partial \bo{v}_{t-1}}{\partial \bo{\alpha}} \right) - \dfrac{\partial \alpha_t}{\partial \bo{\alpha}} \left( \dfrac{\partial \mathcal{L}_{\text{train}}}{\partial \bo{\theta}_{t-1}} + \xi_t \bo{\theta}_{t-1} \right) \\
& =  -\beta_t \alpha_t \dfrac{\partial \bo{v}_{t-1}}{\partial \bo{\alpha}} - \delta_t^{\otimes}\left(\beta_t \bo{v}_{t-1} + \dfrac{\partial \mathcal{L}_{\text{train}}}{\partial \bo{\theta}_{t-1}} + \xi_t \bo{\theta}_{t-1}\right) 
\end{align*}

Now since 

\begin{equation*}
\bo{v}_{t} = \beta_{t} \bo{v}_{t-1} + \dfrac{\partial \mathcal{L}_{\text{train}}}{\bo{\theta}_{t-1}} + \xi_{t} \bo{\theta}_{t-1}
\end{equation*}

we can write the partial derivative of the velocity as an another recursive rule:

\begin{align*}
\bo{C}_t^{\bo{\alpha}} &= \dfrac{\partial \bo{v}_t }{\partial \bo{\alpha}} \\
&= \beta_t \bo{C}_{t-1}^{\bo{\alpha}} + \dfrac{\partial^2 \mathcal{L}_{\text{train}}} {\partial \bo{\alpha} \partial \bo{\theta}_{t-1}} + \xi_t \dfrac{\partial \bo{\theta}_{t-1}}{\partial \bo{\alpha}} \\
& = \beta_t \bo{C}_{t-1}^{\bo{\alpha}} + \left( \xi_t \vone + \dfrac{\partial^2 \mathcal{L}_{\text{train}}}{\partial \bo{\theta}_{t-1}^2} \right) \dfrac{\partial \bo{\theta}_{t-1}}{\partial \bo{\alpha}}
\end{align*}

And putting all together recovers the system:

\begin{equation*}
\systeme{
\bo{A}_t^{\bo{\alpha}} = \vone  - \alpha_t \left(\dfrac{\partial^2 \mathcal{L}_{\text{train}}}{\partial \bo{\theta}_{t-1}^2} + \xi_t \vone \right),
\bo{B}_t^{\bo{\alpha}} = -\beta_t \alpha_t \bo{C}_{t-1}^{\bo{\alpha}} - \delta_t^{\otimes}\left(\beta_t \bo{v}_{t-1} + \dfrac{\partial \mathcal{L}_{\text{train}}}{\partial \bo{\theta}_{t-1}} + \xi_t \bo{\theta}_{t-1}\right),
\bo{C}_t^{\bo{\alpha}} = \beta_t \bo{C}_{t-1}^{\bo{\alpha}} + \left(\xi_t \vone  + \dfrac{\partial^2 \mathcal{L}_{\text{train}}}{\partial \bo{\theta}_{t-1}^2} \right) \bo{Z}_{t-1}^{\bo{\alpha}}
}
\end{equation*}

For learning the momentum and weight decay, a very similar approach yields

\begin{equation*}
\systeme{
\bo{A}_t^{\bo{\beta}} = \vone  - \alpha_t \left(\dfrac{\partial^2 \mathcal{L}_{\text{train}}}{\partial \bo{\theta}_{t-1}^2} + \xi_t \vone \right),
\bo{B}_t^{\bo{\beta}} = -\beta_t \alpha_t \bo{C}_{t-1}^{\bo{\beta}} - \delta_t^{\otimes}(\alpha_t \bo{v}_{t-1}),
\bo{C}_t^{\bo{\beta}} = \delta_t^{\otimes}(\bo{v}_t) + \beta_t \bo{C}_{t-1}^{\bo{\beta}} + \left(\xi_t \vone  + \dfrac{\partial^2 \mathcal{L}_{\text{train}}}{\partial \bo{\theta}_{t-1}^2} \right) \bo{Z}_{t-1}^{\bo{\beta}}
}
\end{equation*}

and

\begin{equation*}
\systeme{
\bo{A}_t^{\bo{\xi}} = \vone  - \alpha_t \left(\dfrac{\partial^2 \mathcal{L}_{\text{train}}}{\partial \bo{\theta}_{t-1}^2} + \xi_t \vone \right),
\bo{B}_t^{\bo{\xi}} = -\beta_t \alpha_t \bo{C}_{t-1}^{\bo{\xi}} - \delta_t^{\otimes}(\alpha_t \bo{\theta}_{t-1}),
\bo{C}_t^{\bo{\xi}} = \delta_t^{\otimes}(\bo{\theta}_{t-1}) + \beta_t \bo{C}_{t-1}^{\bo{\xi}} + \left(\xi_t \vone  + \dfrac{\partial^2 \mathcal{L}_{\text{train}}}{\partial \bo{\theta}_{t-1}^2} \right) \bo{Z}_{t-1}^{\bo{\xi}}
}
\end{equation*}

% Finally, we note that strictly speaking sharing hyperparameters over inner steps corresponds to summing the hypergradient from each batch. In order to be able to use the same outer optimizer regardless of the training budget, we thus need to divide hypergradients with the number of steps per hyperparameter to get an average.

\clearpage
\subsection*{\fontsize{12}{17}\selectfont Appendix B: Theorem \ref{theorem} Proof}
\vspace{0.5cm}

\paragraph{Preamble} Consider that each time step $t \in \{1, 2, ..., T\}$ has a corresponding hyperparameter $\lambda_t$ and hypergradient $g_t=\partial L_{\text{val}}(\vtheta_T)/ \partial \lambda_t$. Each $g_t$ is viewed as a random variable due to the sampling process of the weight initialization $\vtheta_0$ and the inner loop minibatch selection. Assume that $\vg = [g_1,g_2,\ldots, g_{T}]$ is sufficiently well approximated by a Gaussian distribution, where $\vg\sim \mathcal{N}(\vmu,\mSigma)$, with mean $\vmu=[\mu_1, \mu_2, ..., \mu_T]$ and covariance matrix $\mSigma$. Assume that the values of $\vmu$ can be written as the $\epsilon$-Lipschitz function $\mu_{t+1} = \mu_t + \epsilon_t$, where $\epsilon_t \in [-\epsilon,\epsilon]$. Note that in general, the gradients at different time steps may be correlated. Let the magnitude of the correlation be bounded by $c \in [0,1]$: 
\begin{equation}
\frac{|\Sigma_{tt^\prime}|}{\sqrt{\Sigma_{tt} \Sigma_{t^\prime t^\prime}}} \leq c ~~~~~~~ \forall ~~ t\ne t^\prime
\label{eqn:corrbound}
\end{equation}

Let $W$ define the size of a non-overlapping window over which hypergradients are averaged. This produces $K$ windows, where each window $k\in \{1, 2, ..., K\}$ contains the time steps $S^{(k)}$ i.e. $S^{(1)}=\{1,2,\ldots W\}$, $S^{(2)}=\{W+1,W+2,\ldots 2W\}$, etc. For simplicity of analysis \footnote{This assumption is unnecessary and can be relaxed but would result in a more cumbersome theorem statement, as the final window of size $< W$ would need to be considered.} we assume the chosen window sizes are divisors of $T$ such that $K=T/W$. 
Sharing hyperparameters over $W$ contiguous time steps amounts to using the average hypergradient $\bar{g}^{(k)}$ for each step in that window, where
\begin{equation}
\bar{g}^{(k)} \coloneqq \frac{1}{W} \sum_{t \in S^{(k)}} g_t
\end{equation}

We can now consider the mean squared error across all time steps when averaging contiguous hypergradients in non-overlapping windows of size $W$:
\begin{equation}
\mathrm{MSE}_W = \frac{1}{K}\sum_k \frac{1}{W} \sum_{t \in S^{(k)}} \mathbb{E}\left[\left(\bar{g}^{(k)} -\mu_t\right)^2\right]
 \label{eqn:mse}
\end{equation}
where all expectations in our proof are over $\vg\sim \mathcal{N}(\vmu,\mSigma)$. Note the case $\mathrm{MSE}_1$ gives the standard case where no averaging occurs ($K=T$).

\paragraph{Theorem} Then
\begin{align}
\mathrm{MSE}_1 &= \frac{1}{T} \sum_t \mSigma_{tt} \\
\mathrm{MSE}_W &\leq \frac{(1+c(W-1))}{W}\mathrm{MSE}_1 +\epsilon^2\frac{(W^2-1)}{12}
\end{align}

\paragraph{Proof}
The case for $\mathrm{MSE_1}$ follows trivially from the definition of variance: 
\begin{equation}
\mathrm{MSE}_1 = \frac{1}{T} \sum_t \mathbb{E}[(g_t -\mu_t)^2] = \frac{1}{T} \sum_t \mSigma_{tt} 
\end{equation}

and so the mean squared error is the average of the variances. We now focus on the $W>1$ case. Consider a window enumerated by $k$, and the vector of gradients within that window $\vg^{(k)}=(g_t|t\in S^{(k)})$. Under the Gaussian assumption, that vector is Gaussian distributed with covariance $\mSigma^{(k)}$, which is a block of the covariance matrix $\mSigma$  corresponding to the variables in $\vg^{(k)}$. Let $\bar{\mu}^{(k)}=(1/W)\sum_{t \in S^{(k)}} \mu_t$ be the average of means in window $k$. We consider the mean squared error from window $k$ as :

\begin{align}
\mathrm{MSE}^{(k)} &= \frac{1}{W}\sum_{t \in S^{(k)}} \mathbb{E}\left[\left(\bar{g}^{(k)}-\mu_t\right)^2\right] \\ 
&= \frac{1}{W}\sum_{t \in S^{(k)}} \mathbb{E} \left[\left(\bar{g}^{(k)}-\bar{\mu}^{(k)}\right)^2  + \left(\mu_t - \bar{\mu}^{(k)}\right)^2 - 2\left(\bar{g}^{(k)}-\bar{\mu}^{(k)}\right)\left(\mu_t - \bar{\mu}^{(k)}\right)\right] \\
&= \mathbb{E} \left[ \left(\bar{g}^{(k)}-\bar{\mu}^{(k)}\right)^2 \right]+\frac{1}{W}\sum_{t \in S^{(k)}} \left(\mu_t-\bar{\mu}^{(k)}\right)^2
\label{eqn:MSEk}
\end{align}

Now $\bar{g}^{(k)}=(1/W)\vone^T \vg^{(k)}$, and $\bar{\mu}^{(k)}=(1/W)\vone^T \vmu^{(k)}$, and so $\bar{g}^{(k)}-\bar{\mu}^{(k)}=(1/W)\vone^T (\vg^{(k)}-\vmu^{(k)})$. Hence

\begin{align}
\mathbb{E}\left[\left(\bar{g}^{(k)}-\bar{\mu}^{(k)}\right)^2\right]&=\frac{1}{W^2} \mathbb{E}\left[\vone^T \left(\vg^{(k)} -\vmu^{(k)}\right)\left(\vg^{(k)}-\vmu^{(k)}\right)^T\vone\right] \\ 
&=\frac{1}{W^2}\mathbf{1}^T\mSigma^{(k)}\mathbf{1}
\end{align}

Now let $\mD$ be the diagonal matrix of variances, i.e. $\mD_{ii}=\mSigma^{(k)}_{ii} ~~\forall i$ and $\mD_{ij} = 0 ~~\forall i\neq j$. We use the correlation bound (\ref{eqn:corrbound}), which can be written as $|\mSigma^{(k)}_{ij}|< c \left[ \mD^{\frac{1}{2}}\mathbf{1}\mathbf{1}^T \mD^{\frac{1}{2}}\right]_{ij} ~~ \forall i\neq j$, and this allows us to write an upper bound for the expression above: 

\begin{align}
\mathbb{E}\left[(\bar{g}^{(k)}-\bar{\mu}^{(k)})^2\right]&\le\frac{1}{W^2}\mathbf{1}^T[(1-c)\mD+ c\mD^{\frac{1}{2}} \mathbf{1}\mathbf{1}^T \mD^{\frac{1}{2}}]\mathbf{1}\\
&= \frac{(1-c)}{W^2}\sum_i \mD_{ii} + c \left[\frac{1}{W}\mathbf{1}^T \mD^{\frac{1}{2}}\mathbf{1}\right] \left[\frac{1}{W}\mathbf{1}^T \mD^{\frac{1}{2}}\mathbf{1}\right]\\
&= \frac{(1-c)}{W^2}\sum_i \mD_{ii} + c \left[\frac{1}{W}\sum_i \sqrt{\mD_{ii}}\right]^2 \\
&= \frac{(1-c)}{W^2}\sum_i \mSigma^{(k)}_{ii} + c \left[\frac{1}{W}\sum_i \sqrt{\mSigma^{(k)}_{ii}}\right]^2
\end{align}

This expression can be simplified further using Jensen's inequality for square roots:
\begin{align}
\mathbb{E}\left[(\bar{g}_k-\bar{\mu}_k)^2\right] &\le
 \frac{(1-c)}{W^2} \sum_i \mSigma^{(k)}_{ii} +\frac{cW}{W^2} \sum_i \mSigma^{(k)}_{ii} \\ 
 &= \frac{1+c(W-1)}{W^2}\sum_i \mSigma^{(k)}_{ii} 
 \label{eqn:firsthalf}
 \end{align} 

Now we return to the second part of (\ref{eqn:MSEk}). This second term can be bounded using the Lipschitz constraints. In particular for window size $W$, the maximum error is given when there is maximum deviation from the mean, which occurs when $\mu_t = \mu_{t-1} + \epsilon$. If we write the first mean in window $k$ as $\mu^{(k)}_1$ we have  $\mu_t = \mu^{(k)}_1 + (t-1)\epsilon ~~\forall t \in S^{(k)}$ and in that case $\bar{\mu}^{(k)} = \frac{1}{W}(\mu^{(k)}_1 + (\mu^{(k)}_1+\epsilon) + (\mu^{(k)}_1+2\epsilon) + \ldots + (\mu^{(k)}_1+(W-1)\epsilon) = \mu^{(k)}_1 + \frac{(W-1)\epsilon}{2}$ and so $\mu_t-\bar{\mu}_k = (t-1)\epsilon + \frac{(W-1)\epsilon}{2} $. Note that this quantity is the same for all windows $k$. We can use it to write an upper bound as follows:

\begin{align}
    \frac{1}{W}\sum_{t \in S^{(k)}}\left(\mu_t-\bar{\mu}^{(k)}\right)^2
    &\le \frac{1}{W}\sum_{j=1}^{W} \epsilon^2\left( (j-1) -\frac{W-1}{2} \right)^2 \\
    &=   \frac{\epsilon^2}{W}\sum_{j=0}^{W-1} \left(j -\frac{W-1}{2} \right)^2 \\
    &= \frac{\epsilon^2}{W}\sum_{j=0}^{W-1}j^2-(W-1)j + \frac{(W-1)^2}{4}\\
    &= \frac{\epsilon^2}{W}\left(\frac{W(W-1)(2W-1)}{6}- (W-1)\frac{W(W-1)}{2} + \frac{W(W-1)^2}{4} \right)\\
    &= \epsilon^2\frac{(W^2-1)}{12} \label{eqn:secondhalf}
\end{align}

Hence combining (\ref{eqn:firsthalf}) and (\ref{eqn:secondhalf}) together into (\ref{eqn:MSEk})  we have
\begin{align}
\mathrm{MSE}^{(k)} &\le \frac{1+c(W-1)}{W^2}\sum_i \mSigma^{(k)}_{ii} + \epsilon^2\frac{(W^2-1)}{12}
\end{align}
and so incorporating it into (\ref{eqn:mse}) we get

\begin{align}
\mathrm{MSE}_W &= \frac{1}{K}\sum_{k=1}^{K} \mathrm{MSE}^{(k)}\\
&\le \frac{W}{T} \sum_{k=1}^{K} \left( \frac{(1+c(W-1))}{W^2}\sum_i \mSigma^{(k)}_{ii} +\epsilon^2\frac{(W^2-1)}{12} \right) \\
&= \frac{(1+c(W-1))}{WT}\sum_i \mSigma_{ii} + \epsilon^2\frac{(W^2-1)}{12} \\
&=  \frac{(1+c(W-1))}{W} \mathrm{MSE}_1 + \epsilon^2\frac{(W^2-1)}{12} \myqed
\end{align}

For sufficiently small $\epsilon$ and $c$ we have with certainty, $MSE_W<MSE_1$ for some $W>1$.

\paragraph{Discussion} We assume that means $\vmu$ can be written as the $\epsilon$-Lipschitz function $\mu_{t+1} = \mu_t + \epsilon_t$, where $\epsilon_t \in [-\epsilon,\epsilon]$ . Generally speaking, contiguous hyperparameters have optimal values which are close, and therefore  have close hypergradients during outer optimization. This assumption breaks if hyperparameters are initialized randomly, and so we initialize all of our hyperparameters to zero in our experiments. Ideally, we would solve for whole inner loop several times so that we can use the mean hypergradient $[\mu_0, \mu_1, ..., \mu_H]$ for each individual step, without doing any sharing. While we consider this to be the optimal hypergradients, this is too expensive in practice, and instead we consider averaging hypergradients from neighbouring inner steps. The result above indicates that when contiguous hypergradients are sufficiently de-correlated (small c), we can reduce the mean squared error by a factor $W$ compared to not averaging. However, if means $\mu_t$ drift over time by an amount $\epsilon_t\le\epsilon$ this introduces some bias which increases the error and eventually results in $MSE_W>MSE_1$. 

Finally, it is worth considering the simpler scenario when each hypergradient is considered to be drawn independently, i.e. $\vg\sim \mathcal{N}(\vmu,\sigma^2\mathbf{1})$. In that case, $c=0$ and the mean squared errors become:
\begin{align}
\mathrm{MSE}_1 &= \sigma^2 \\
\mathrm{MSE}_W &\leq \frac{\mathrm{MSE}_1}{W} +\epsilon^2\frac{(W^2-1)}{12}
\end{align}

\paragraph{Visualizing the MSE for various $\mu_t$ profiles}. Since the mean squared error depends on the specific shape of $\mu_t$, we sample random $\mu_t$ profiles and show how their $MSE$ evolves as a function of $W$. This illustrates how tight the upper bound is.
\begin{figure*}[!htbp]
    \centering
    \includegraphics[width=0.9\textwidth]{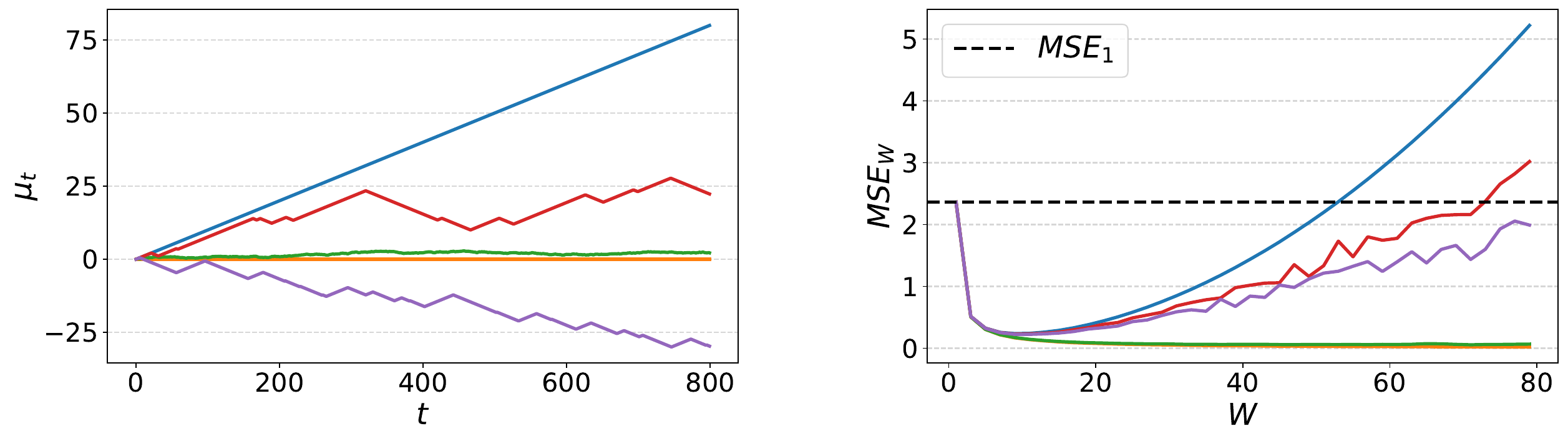}
    \caption{Several $\mu_t$ profiles and their corresponding mean squared error when sharing over $W$ contiguous steps, as a function of $W$.  The blue and yellow curve correspond to the maximal and minimal drifts scenarios respectively.}
    \label{fig:mu_t_profiles}
\end{figure*}

\clearpage
\subsection*{\fontsize{12}{17}\selectfont Appendix C: Implementation Details}
\vspace{0.5cm}

We use a GeForce RTX 2080 Ti GPU for all experiments. We found that much of the literature on greedy methods uses the test set as the validation set, which creates a risk of meta-overfitting to the test set. Instead, we always carve out a validation set from our training set.

\paragraph{Figure 1} Here we used very similar settings as Figure 4 for FDS, except we learned $7$ learning rates to make the search space a bit more challenging. We use the HyperBanster HPO package for RS, BO, HB and BOHB. For HB and BOHB, the minimum budget argument is set to 1 epoch to allow for lots of fast evaluations, and the maximum budget is set to $50$ epochs. We also use this technique to bring down our convergence time from $\sim 10$ hours to $\sim 3$ hours, namely we calculate hypergradients based on 10 epochs for some outer steps, rather than calculate all hypergradients on 50 epochs. Since HD needs the user to specify initial hyperparameter values, we random search over those for several consecutive HD runs. 

\paragraph{Figure 2} We calculate the hypergradient with respect to some learning rate schedule over $100$ seeds, where each seed corresponds to a different training set ordering and network initialization. The learning rate schedule is fixed, and initialized to be a cosine decay over the full $250$ batches, starting at $0.01$. The batch size is set to $128$, and $1000$ fixed images are used for the validation data.

\paragraph{Figure 3} Here we used a batch size of $128$ for both datasets to allow $1$ epoch worth of inner optimization in about $500$ inner steps. Clipping was restricted to $\pm 3$ to show the effect of noisy hypergradients more clearly. Since MNIST and SVHN are cheap datasets to run on a LeNet architecture, we can afford $50$ outer steps and early stopping based on validation accuracy. All learning rates were initialized to zero.

\paragraph{Figure 4} We learn $5$ values for the learning rates, $1$ for the momentum and $1$ for the weight decay, to make it comparable to the hyperparameters used in the literature for CIFAR-10. A batch size $256$ is used, with $5\%$ of the training set of each epoch set aside for validation. We found larger validation sizes not to be helpful. Hypergradient descent uses hyperparameters initialized at zero as well, and trains all hyperparameters online with an SGD outer optimizer with learning rate $0.2$ and $\pm 1$ clipping of the hypergradients. As described in appendix G, we used a sign based outer optimizer with adaptive step sizes rather than some hand-tuned outer learning rate schedule. We used initial values $\gamma_\alpha = 0.1, \gamma_\beta = 0.15$ and $\gamma_\xi = 4\times 10^{-4}$ but the performance barely changed when these values were multiplied or divided by 2. Since we take $10$ outer steps and initialize all hyperparameters at zero, this defines a search ranges: $\alpha \in [-1,1]$, $\beta \in [-1.5, 1.5]$, and $\gamma \in [-4\times10^{-3}, 4\times10^{-3}]$. The Hessian matrix product is clipped to $\pm 10$ to prevent one batch from having a dominating contribution to hypergradients.

\clearpage
\subsection*{\fontsize{12}{17}\selectfont Appendix D: Other hypergradient examples}
\vspace{0.5cm}

Figure \ref{fig:hypergrads_and_MSE} depends on the value of $\bo{\alpha}$ at which hypergradients are calculated. For some learning rate schedules, contiguous hypergradients are sampled from closer distribution ($\epsilon$ small) and so sharing over larger windows is beneficial, as shown in the figure below. 

\begin{figure*}[!htbp]
    \centering
    \includegraphics[width=\textwidth]{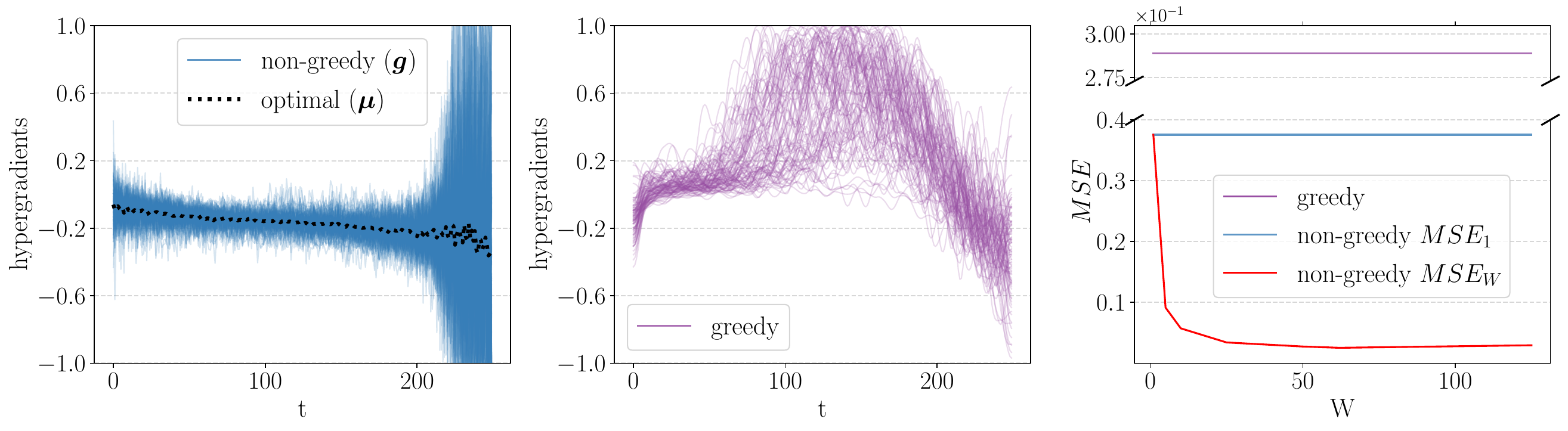}
    \caption{ Similar to Figure \ref{fig:hypergrads_and_MSE} but  for a smaller $\epsilon$. We can see that averaging hypergradients helps even more.}
    \label{fig:hypergrads_and_MSE_v2}
\end{figure*}

\subsection*{\fontsize{12}{17}\selectfont Appendix E: Hypergradients}
\vspace{0.5cm}

Here we provide the raw hypergradients corresponding to the outer optimization shown in Appendices: Figure 1. Note that the range of these hypergradients is made reasonable by the averaging of gradients coming from contiguous hyperparameters.

\begin{figure*}[!htbp]
    \centering
    \includegraphics[width=1.0\textwidth]{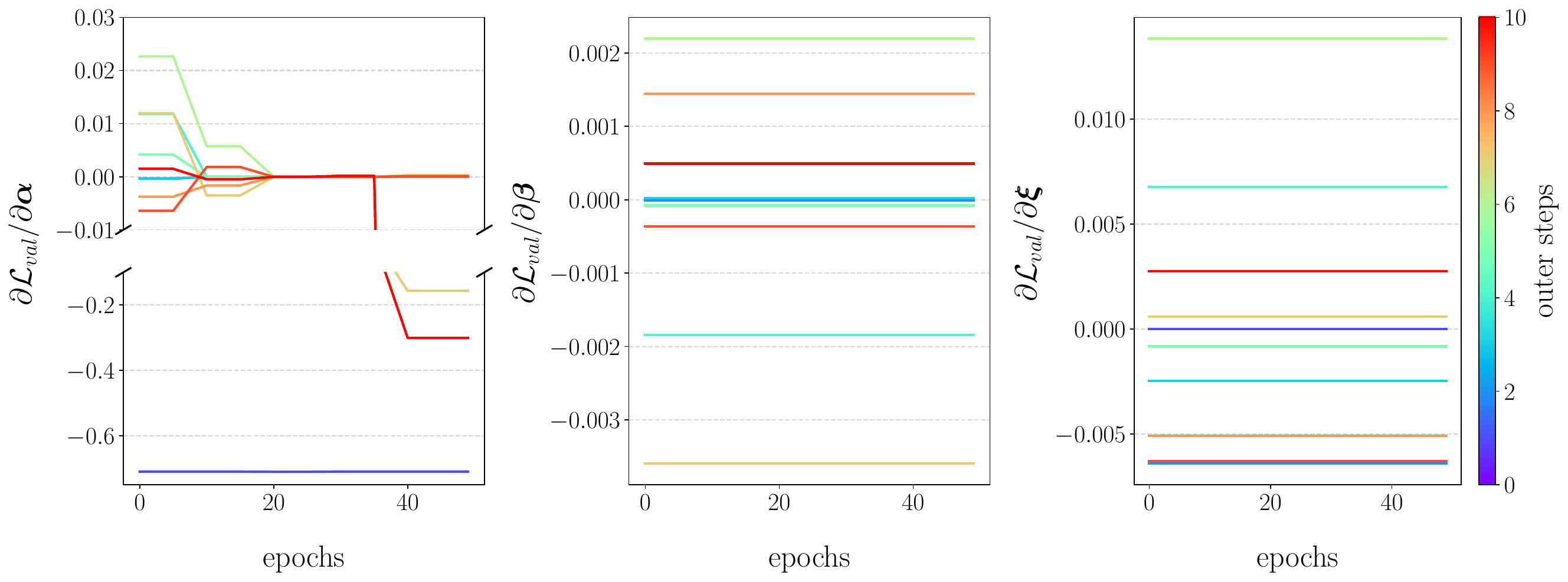}
    \caption{Hypergradients have a reasonable range but fail to always converge to zero when the validation performance stops improving.}
    \label{fig:hypergradients}
\end{figure*}

\clearpage
\subsection*{\fontsize{12}{17}\selectfont Appendix F: Baselines}
\vspace{0.5cm}

\begin{figure*}[!htbp]
    \centering
    \includegraphics[width=\textwidth]{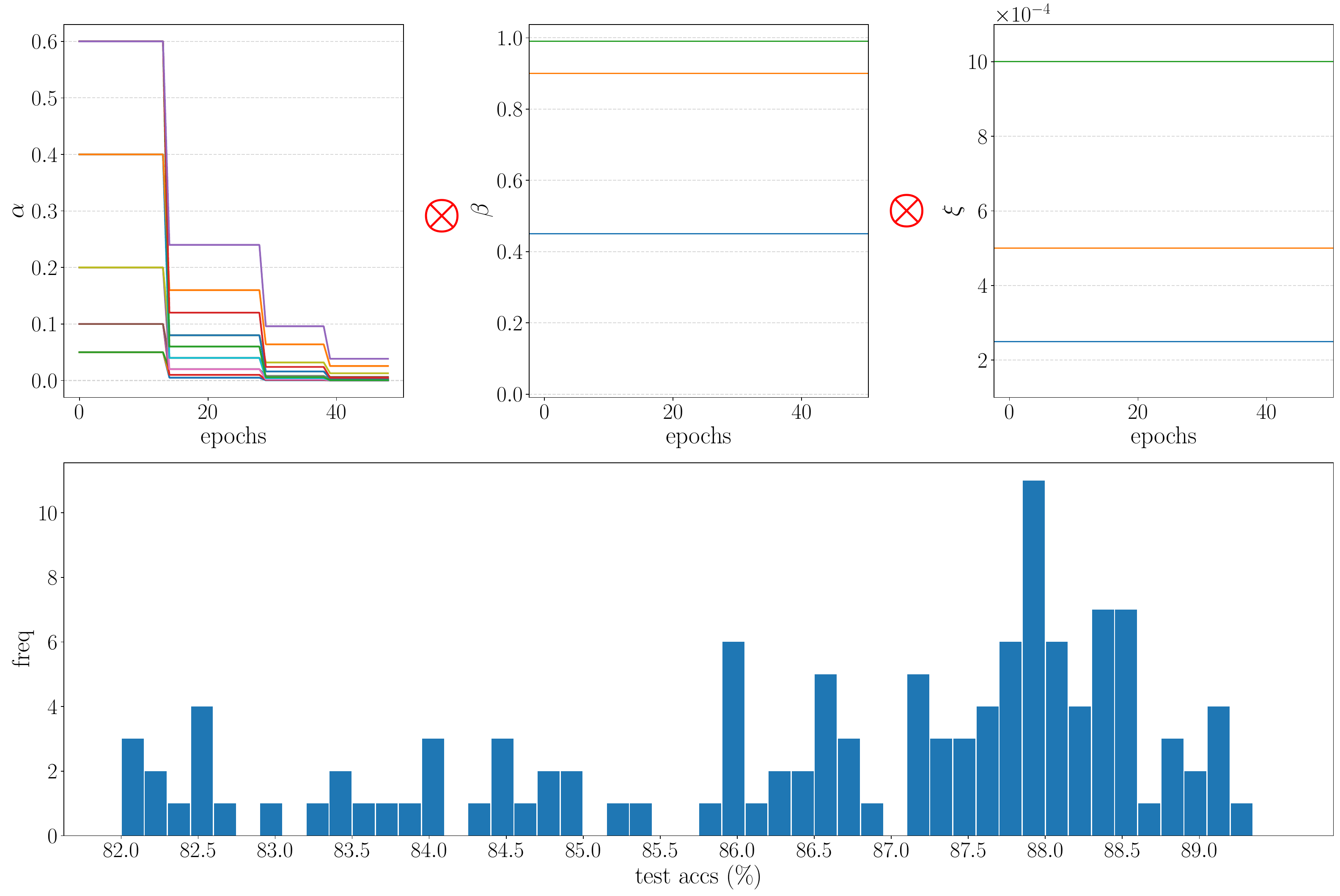}
    \caption{The combination of hyperparameters searched over for CIFAR-10 (top row) and the corresponding distribution of test accuracies (bottom row).}
    \label{fig:baselines}
\end{figure*}

The objective here is to select the best hyperparameter setting that a deep learning practitioner would reasonably be expected to use in our experimental setting, based on the hyperparameters used by the community for the datasets at hand. For CIFAR-10, the most common hyperparameter setting is the following: $\alpha$ is initialized at $\alpha_0 = 0.2$ (for batch size $256$, as used in our experiments) and decayed by a factor $\eta=0.2$ at $30\%, 60\%$ and $80\%$ of the run (\texttt{MultiStep} in Pytorch); the momentum $\beta$ is constant at $0.9$, and the weight decay $\xi$ is constant at $5 \times 10^{-4}$. We search for combinations of hyperparameters around this setting. More specifically, we search over all combinations of $\alpha_0 = \{0.05, 0.1, 0.2, 0.4, 0.6\}$, $\eta = \{0.1, 0.2, 0.4\}$, $\beta = \{0.45, 0.9, 0.99\}$, and $\xi = \{2.5\times10^{-4}, 5\times10^{-4}, 1\times10^{-3} \}$. This makes up a total of $135$ hyperparameter settings, which we each run $3$ times to get a mean and standard deviation. The distribution of those means are provided in Figure \ref{fig:baselines}, and the best hyperparameter setting is picked based on validation performance, which corresponds to $89.2 \pm 0.2\%$. Preliminary experiments showed that using schedules for the momentum and weight decay did not improve test accuracy.

\newpage

\clearpage
\subsection*{\fontsize{12}{17}\selectfont Appendix G: Using Hypergradient Signs}
\vspace{0.5cm}
While hyperparameter sharing produces stable hypergradients with a reasonable range (see Appendix E), tuning the outer learning rate schedule can be tedious and unintuitive, and doesn't allow the user to specify a range to search hyperparameters over. A simple and fairly common learning rate schedule consists in decaying the learning rate (e.g. by a factor of 2) every time the gradient changes sign \cite{Jacobs1988DecayLearningRateFromSign}. The idea of using the sign of gradients to improve the efficiency of gradient descent dates back to the RPROP optimizer \cite{Riedmiller1993RPROP}. More recently, gradient signs have also been used to improve efficiency in the context of distributed learning \cite{Bernstein2018SignSGD}, which has led to the discovery of robust convergence properties of sign-based SGD even in the case of biased gradients \cite{Safaryan2019StochasticSignDescentMethods}. Using such a learning rate schedule for the outer optimizer frees us from having to tune the outer learning rate, but fails to define a search range for HPO. We can achieve this by updating each hyperparameter by an amount $\sgn (g) \times \gamma$, and letting $\gamma \leftarrow \gamma/2$ when the hypergradient $g$ changes sign across 2 consecutive outer steps. This allows for convergence after hypergradients have changed signs a few times. Being able to define the range of hyperparameter search more explicitly is especially useful to compare FDS to other HPO algorithms which also use a fixed search range (section \ref{sec:exp_HPO_baselines}).

\vspace{1cm}

\begin{wrapfigure}[26]{R}{0.5\textwidth}
\begin{minipage}[b]{0.5\textwidth}
\vspace{-0.85cm}
\begin{algorithm}[H]

     \caption{FDS algorithm similar to algorithm \ref{alg:forward_mode} but using hypergradient signs to update hyperparameters.}
     \label{alg:forward_mode_with_sign}

\begin{algorithmic}
\STATE {\bfseries Initialize:} $ N^\alpha, W = T/N^\alpha,\bo{\alpha} = \bo{0}^{N^\alpha}$,\\ $\bo{\gamma} \in \mathbb{R}^{N^\alpha}$
\vspace{.22cm}

\STATE {\color{CadetBlue} \textit{\#outer loop}}
 \FOR{$o$ in $1, 2, ...$} 
        \STATE {\bfseries Initialize:} $\mathcal{D}_{\text{train}}$, $\mathcal{D}_{\text{val}}$, $\bo{\theta}_0 \in \mathbb{R}^D$,  \\$\mZ^{\bo{\alpha}} = \bo{0}^{D \times N^\alpha}$, $C^{\bo{\alpha}} = \bo{0}^{D \times N^\alpha}$

        \vspace{.22cm}
        \STATE {\color{CadetBlue}\textit{\#inner loop}}
        \FOR{$t$ in $1, 2, ..., T$}
              \STATE $\bo{x}_{\text{train}}, \bo{y}_{\text{train}} \sim \mathcal{D}_{\text{train}}$
              \STATE $\bo{g}_{\text{train}} = \partial\mathcal{L}_{\text{train}}(\bo{x}_{\text{train}}, \bo{y}_{\text{train}}) / \partial \bo{\theta}$
              
              \vspace{.20cm}
              \STATE {\color{CadetBlue} \textit{\#hyperparameter sharing}}
              \STATE $i = \ceil{t / W}$
              \STATE $\bo{\mathcal{H}}\mZ^{\bo{\alpha}}_{[1:i]} = \partial(\bo{g}_{\text{train}}  \mZ^{\bo{\alpha}}_{[1:i]}) / \partial \bo{\theta}$
              \STATE $\mZ^{\bo{\alpha}}_{[1:i]} = \mA^{\bo{\alpha}}\mZ^{\bo{\alpha}}_{[1:i]} + \mB^{\bo{\alpha}}_{[1:i]}$
              
              \vspace{.25cm}
              \STATE update $C^{\bo{\alpha}}$  (Eq \ref{eq:recursive_alpha})
              \STATE $\bo{\theta}_{t+1} = \Phi(\bo{\theta}_{t}, \bo{g}_{\text{train}})$ 
    \vspace{.25cm}
    \ENDFOR
    
    \vspace{.25cm}
    \STATE $\bo{g}_{\text{val}} = \partial\mathcal{L}_{\text{val}}(\mathcal{D}_{\text{val}}) / \partial \bo{\theta}$
    \STATE $\bo{s}_o = \sgn (\bo{g}_{\text{val}} \mZ^{\bo{\alpha}}/W)$
    
    \vspace{.25cm}
    \FOR{n in $1, 2, ..., N_\alpha$}
		\IF{$\bo{s}_{o, [n]} \neq \bo{s}_{o-1, [n]}$}
	    \vspace{.15cm}
		    \STATE $\bo{\gamma}^{\bo{\alpha}}_{[n]} \leftarrow \bo{\gamma}^{\bo{\alpha}}_{[n]} /2$
	    \vspace{.15cm}
        \ENDIF
    \ENDFOR
        		
    \vspace{.25cm}
    \STATE $\bo{\alpha} \leftarrow \bo{\alpha} - \bo{s}_o \odot \bo{\gamma}^{\bo{\alpha}}$
    \vspace{.25cm}

\vspace{.25cm}
\ENDFOR

\end{algorithmic}
\end{algorithm}
\end{minipage}
\end{wrapfigure}

\end{document}